\pdfoutput=1

\documentclass[11pt]{article}

\usepackage[]{latex/acl}

\usepackage{times}
\usepackage{latexsym}

\usepackage[T1]{fontenc}

\usepackage[utf8]{inputenc}

\usepackage{microtype}

\usepackage{inconsolata}
\usepackage{adjustbox}
\usepackage{amsmath}
\usepackage{amssymb}
\usepackage{dsfont}
\usepackage{subfigure}
\usepackage{graphicx}
\usepackage{CJKutf8}
\usepackage{algorithm}
\usepackage{algpseudocode}
\usepackage{caption}
\usepackage{booktabs}
\usepackage{siunitx}
\usepackage{wrapfig}
\usepackage{rotating}
\usepackage{float}
\usepackage{bm}
\usepackage{hyperref}
\usepackage{multirow}
\usepackage{tabu}
\title{On the Hallucination in Simultaneous Machine Translation}

\author{
    Meizhi Zhong\textsuperscript{\rm 1},\space\space
    Kehai Chen\textsuperscript{\rm 1} \thanks{ Corresponding authors}, \space\space
    Zhengshan Xue\textsuperscript{\rm 2},\space\space
    Lemao Liu\textsuperscript{\rm },\space\space
    Mingming Yang\textsuperscript{\rm },\space\space 
    Min Zhang\textsuperscript{\rm 1}\space\space \\
    \textsuperscript{\rm 1}Institute of Computing and Intelligence, Harbin Institute of Technology, Shenzhen, China \\
    \textsuperscript{\rm 2}College of Intelligence and Computing, Tianjin University, Tianjin, China \\
    {\tt 22s051052@stu.hit.edu.cn},\space\space
    {\tt chenkehai@hit.edu.cn},\space\space
    {\tt xuezhengshan@tju.edu.cn},\\
    {\tt lemaoliu@gmail.com},\space\space
    {\tt shanemmyang@gmail.com},\space\space
    {\tt zhangmin2021@hit.edu.cn}
}

\begin{document}
\begin{CJK}{UTF8}{gbsn}
\maketitle
\begin{abstract}
It is widely known that hallucination is a critical issue in Simultaneous Machine Translation (SiMT) due to the absence of source-side information. 
While many efforts have been made to enhance performance for SiMT, few of them attempt to understand and analyze hallucination in SiMT.
Therefore, we conduct a comprehensive analysis of hallucination in SiMT from two perspectives: understanding the distribution of hallucination words and the target-side context usage of them.
Intensive experiments demonstrate some valuable findings and particularly show that it is possible to alleviate hallucination by decreasing the over usage of target-side information for SiMT. 
~\footnote{Code is available at \url{https://github.com/zhongmz/SiMT-Hallucination}}

\end{abstract}

\section{Introduction}

In neural machine translation, hallucination occurrences are not common due to its small quantity~\cite{lee_hallucinations,yan2022probing,raunak_curious_2021,guerreiro2022looking}.
But in simultaneous machine translation (SiMT), it has been found that  
hallucination is extremely severe, especially as latency increases 
indicating that hallucination is a critical issue in SiMT.
Currently, most prior works concentrate on how to enhance model performance for SiMT~\citep{ma2018stacl,ma2019monotonic,zheng_simultaneous_2020,zhang_information-transport-based_2022,zhang2022modeling,guo2022turning,zhang2022reducing}, however, only a few of them measure the hallucination phenomenon~\cite{chen_improving_2021,deng_improving_2022,liu2023cbsimt}. To our best knowledge, there are no researches which {\em systematically analyze hallucination in SiMT}. 

Therefore, we conduct a comprehensive analysis of hallucinations in SiMT.
Initially, we seek to empirically analyze these hallucination words from the perspective of their distribution. 
We collect all hallucination words together and understand their frequency distribution, and we find that these words are randomly distributed with a high entropy: their entropy is almost as high as that for all target words. 
In addition, to delve into the contextual aspects of hallucination~\cite{xiao2021hallucination}, we consider their predictive distribution.
We discover that their uncertainty is significantly higher than that of non-hallucination words. Furthermore, we find that the SiMT model does not fit the training data well for hallucination words due to the essence of SiMT (i.e., the limited source context), which explains why making correct predictions for hallucination words is difficult.

Intuitively, since a SiMT model is defined on top of a limited source context, this may indirectly cause the model to focus more on the target context and lead to the emergence of hallucination words. 
To verify this intuition, we propose to analyze the usage of the target context for hallucination words for SiMT. Specifically, 
following~\citet{li_word_2019,miao_prevent_2021,fernandes2021measuring,voita_analyzing_2021,yu-etal-2023-promptst,guerreiro2022looking}, 
we firstly employ a metric to measure how much target-context information is used by SiMT with respect to the source-context information. 
With the help of this metric, we find that hallucination is indeed significantly more severe when the SiMT model focuses more on target-side information. 
Drawing upon this, we reduce the over-target-reliance effects by introducing noise into the target-side context.
Experimental results show that the proposed method achieves some modest improvements in terms of BLEU and hallucination effect when the latency is relatively small. 
This discovery gives us some inspiration: more flexible control over the use of target-side information may be a promising approach to alleviate the issue of hallucination.

Our key contributions are as follows:
\begin{itemize}
    \item We study hallucination words from frequency and predictive distributions and observe that the frequency distribution of hallucination words is with high entropy and hallucination words are difficult to be memorized by the predictive distribution during training. 
    
    \item We analyze hallucination words according to the usage of (limited) source context. We find that hallucination words make use of more target-context information than source-context information, and it is possible to alleviate hallucination by decreasing the usage of the target context. 
\end{itemize}

\section{Experimental Settings}
Our analysis is based on the most widely used SiMT models and datasets. This section introduces these models and datasets as follows.
\paragraph{SiMT Models and Datasets.}
SiMT models translate by reading partial source sentences.
\citet{ma2018stacl} proposed widely used Wait-$k$ models for SiMT. It involves reading k words initially and then iteratively generating each word until the end of the sentence. We conducted experiments on it.
We use two standard benchmarks from IWSLT14 De$\leftrightarrow$En~\cite{Cettolo14iwslt} and MuST-C Release V2.0 Zh$\rightarrow$En~\cite{cattoni2021must} to conduct experiments.
Appendix~\ref{detailed exp} provides detailed settings. 
Due to space limitation, we only present the experimental results for the De$\rightarrow$En benchmark. 
The results for Zh$\rightarrow$En and En$\rightarrow$De are similar, as shown in Appendix~\ref{mustczhen} and \ref{iwslt14ende}.

\paragraph{Hallucination Metric.}
In SiMT, \citet{chen_improving_2021} pioneers the definition of Hallucination Metrics based on word alignment $a$. A target word ${\hat{y}}_t$, is a hallucination if there is no alignment to any source word $x_j$. 
This is formally represented as:
\begin{equation}
H(t, a) = \mathds{1} \left[ \{(i, t) \in a \} = \varnothing \right].
\end{equation}
Conversely, a target word ${\hat{y}}_t$, is not a hallucination if there is alignment to any source word $x_j$. 

The Hallucination Rate (\textrm{HR}) is defined as following:
\begin{equation}
\textrm{HR}(x, \hat{y}, a)=\frac{1}{|\hat{y}|} \sum_{t=1}^{|\hat{y}|} H(t, a).
\label{eq:HR}
\end{equation}
\citet{deng_improving_2022} propose GHall to measure hallucination in Wait-$k$. 
Formally, a word is a hallucination if it does not align with the current source:
\begin{equation}
H_{wait-k}(t, a)=\mathds{1}[\{(s, t) \in a \mid s \geq t+k\}=\varnothing].
\end{equation}
The definition of \textrm{HR} remains consistent with ~\citet{chen_improving_2021}.
We utilize GHall metrics to conduct experiments.
We use Awesome-align~\cite{dou_word_2021} as the word aligner $a$.

\section{Understanding Hallucination Words from Distribution}
\begin{table}[t]
  \centering
  \begin{adjustbox}{width=\columnwidth}
  \begin{tabular}{c|cccccc}
    \toprule
    \textbf{$k$} & $1$ & $3$ & $5$ & $7$ & $9$ & $\infty$ \\
    \midrule
    \textrm{HR} $\%$ &{31.28} & {22.57} & {18.58} & {16.41} & {15.21} & 11.50 \\
    \bottomrule
  \end{tabular}
  \end{adjustbox}
  \caption{\textrm{HR} on valid set of wait-$k$, where $k=\infty$ means Full-sentence MT.}
  \label{tab:dhr_in_waitk}
\end{table}
\paragraph{Hallucination is severe in SiMT.}
We measure \textrm{HR} of Wait-$k$ models, illustrated in Table~\ref{tab:dhr_in_waitk}. 
We obtain that Wait-$k$ models suffer more from hallucinations than Full-sentence MT. 
Furthermore, with $k$ decreasing, hallucinations increase clearly.
This shows that hallucination is an important issue and it is worth the in-depth study.

\subsection{Understanding Hallucination from Frequency Distribution}

\begin{figure}[t]
  \centering
  \includegraphics[width=2.5in]{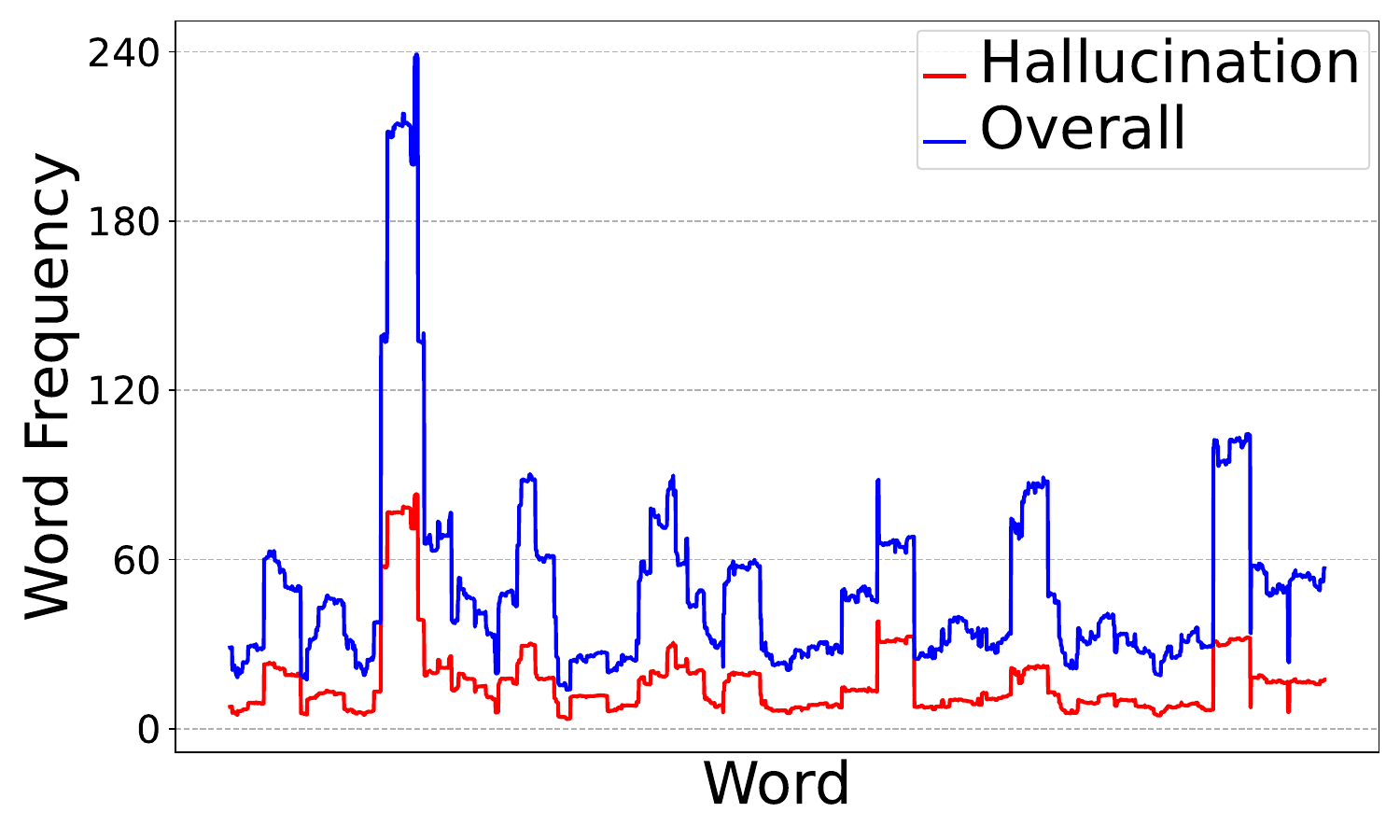}
  \captionof{figure}{Word frequency of Hallucination and Overall on valid hypotheses set of wait-$1$ (x-axis is ordered randomly, with additional $k$ results in Appendix~\ref{Distribution}).}
  \label{fig:word_freq_hall_vs_overall_wait1}
\end{figure}

\begin{table}[t]
  \centering
  \begin{adjustbox}{width=0.9\columnwidth}
  \begin{tabular}{c|cccccc}
    \toprule
    \textbf{$k$} & $1$ & $3$ & $5$ & $7$ & $9$ \\
    \midrule
    Hallucination & 7.82 & 8.22 & 8.19 & 8.10 & 8.07  \\
    Overall & 8.70 & 8.97 & 9.00 & 9.01 & 9.02 \\
    \bottomrule
  \end{tabular}
  \end{adjustbox}
  \caption{Word frequency distribution entropy of Hallucination and Overall on the valid set of wait-$k$.}
  \label{tab.entropy}
\end{table}
\paragraph{Hallucination words are with high distribution entropy.}
To investigate hallucination words in Wait-$k$, we compare frequency distributions of hallucination and overall words.
Figure~\ref{fig:word_freq_hall_vs_overall_wait1} and Table~\ref{tab.entropy} illustrate that their distributions are remarkably similar and both exhibit high entropy. 
It suggests that understanding hallucination from high distribution entropy is challenging.



\subsection{Understanding Hallucination from Predictive Distribution}
We investigate \textbf{Confidence} and \textbf{Uncertainty} of the predictive distribution. 
We define the Confidence of a word as its probability and the Uncertainty of a word as the entropy of its predictive distribution.

\begin{table}[t]
\centering
\resizebox{1.0\columnwidth}{!}{
\begin{tabular}{lcccccccc}
\toprule
\multirow{4}{*}{\bf Wait-$k$}                    & \multicolumn{4}{c}{\bf Valid set} & \multicolumn{4}{c}{\bf Training subset} \\ 
\cmidrule(lr){2-5}\cmidrule(lr){6-9}
&  \multicolumn{2}{c}{\bf Uncertainty}        &  \multicolumn{2}{c}{\bf Confidence}       &  \multicolumn{2}{c}{\bf Uncertainty}        &  \multicolumn{2}{c}{\bf Confidence}       \\ 
\cmidrule(lr){2-3}\cmidrule(lr){4-5}\cmidrule(lr){6-7}\cmidrule(lr){8-9}
&\bf  H        &\bf  NH       &\bf  H        &\bf  NH   & \bf  H        &\bf  NH       &\bf  H        &\bf  NH       \\ 
\midrule
\multicolumn{1}{c}{$k$=1}     & 3.53         & 2.35       & 0.40         & 0.61     & 3.47                & 2.13 & 0.41         & 0.65   \\ 
\multicolumn{1}{c}{$k$=3}     & 3.00         & 2.04       & 0.48         & 0.66     & 2.98                & 1.90 & 0.49         & 0.69   \\ 
\multicolumn{1}{c}{$k$=5}     & 2.81         & 1.97       & 0.52         & 0.67     & 2.76                & 1.90 & 0.52         & 0.69   \\ 
\multicolumn{1}{c}{$k$=7}     & 2.55         & 1.89       & 0.55         & 0.69    & 2.48                & 1.81 & 0.57         & 0.70    \\ 
\multicolumn{1}{c}{$k$=9}     & 2.48         & 1.92       & 0.57         & 0.68     & 2.42                & 1.96 & 0.58         & 0.69   \\ 
\bottomrule
\end{tabular}}
\caption{The Uncertainty and Confidence of Hallucination ({\bf H}) and Non-Hallucination ({\bf NH}) on the valid set and training subset of wait-$k$ models.}
\label{tab:Unc_Con_in_valid_train}
\end{table}

\paragraph{Hallucination words are difficult to translate.}
To explore the difficulty of translating hallucination and non-hallucination words, we calculate the average confidence and uncertainty on the valid set.
The results in the left of Table~\ref{tab:Unc_Con_in_valid_train} reveal that during decoding hallucination words, the models exhibit higher uncertainty. Additionally, the confidence is lower. 
It suggests that models encounter challenges in accurately translating hallucination words.

\paragraph{Hallucination words are difficult to memorize.}
To investigate the reasons behind the difficulty in translating hallucination words, we measure confidence and uncertainty for hallucination and non-hallucination words on the training data. We sample examples from the training data as a training subset with the same size as the valid set.
The results in the right of Table~\ref{tab:Unc_Con_in_valid_train} illustrate that even in previously encountered contexts, models remain uncertain when dealing with hallucination words. 
These findings suggest that models do not fit well with hallucination words during training, leading to a limited ability to generalize to similar contexts on the valid set. 
Consequently, the difficulty in translating hallucination words can be attributed to challenges in memorization during the training.
Additionally, we observe that as $k$ increases, the uncertainty decreases significantly. 
It can be attributed to the model encountering source-side context more, enabling a improved
memorization.

\section{Analysis of Target Context Usage for Hallucination Words}
To verify the hypothesis that using more on target-side context leads to the emergence of hallucination, we propose to analyze the usage of target-side context.

\paragraph{Measure on Target-side Context Usage.}
To explicitly measure Target Context Usage, we adapt an interpretive approach that evaluates the relevance of both target and source words. It involves deactivating connections between the corresponding words and the network. We compute the relevance between the words in the source or target and the next word to be generated and determine the maximum absolute relevance as source or target relevance. It allows us to calculate the \textbf{T}arget-Side Relevance to \textbf{S}ource-Side Relevance '\textbf{s} \textbf{R}atio (TSSR). 

To begin with, we assess the relevance of target-side words and source-side words to the next word to be generated. This evaluation is conducted by selectively deactivating the connection between $\mathrm{x}_j$ or $\mathrm{y}_j$ and the encoder or decoder network in a deterministic manner, following the approach described in \citet{li_word_2019}. 
More formally, the relevance $R(\mathrm{y}_i, \mathrm{x}_j)$ or $R(\mathrm{y}_i, \mathrm{y}_j)$ in Wait-$k$ is directly determined through the dropout effect on $\mathrm{x}_j$ or $\mathrm{y}_j$, as outlined below:
\begin{align}
R\left(\mathrm{y}_i, \mathrm{x}_j\right) &= P\left(\mathrm{y}_i \mid \mathbf{y}_{<i}, \mathbf{x}_{\leq i+k-1}\right) \nonumber \\
&\quad - P\left(\mathrm{y}_i \mid \mathbf{y}_{<i}, \mathrm{x}_{\leq i+k-1,(j, \mathbf{0})}\right).
\label{eq:src_token_relevance}
\end{align}

\begin{align}
R\left(\mathrm{y}_i, \mathrm{y}_j\right) &= P\left(\mathrm{y}_i \mid \mathbf{y}_{<i}, \mathbf{x}_{\leq i+k-1}\right) \nonumber \\
&\quad - P\left(\mathrm{y}_i \mid \mathbf{y}_{<i,(j, \mathbf{0})}, \mathrm{x}_{\leq i+k-1}\right).
\label{eq:tgt_token_relevance}
\end{align}

The relevance of the source-side and target-side is determined by selecting the maximum absolute value of the word's relevance on the current source-side and the current target-side. Formally, this can be expressed as:
\begin{equation}
R\left(\mathrm{y}_i\right)_{source-side}=\max\{ \lvert R\left(\mathrm{y}_i, \mathrm{x}_j\right) \rvert \}.
\label{eq:src_sentence_relevance}
\end{equation}

\begin{equation}
R\left(\mathrm{y}_i\right)_{target-side}=\max\{ \lvert R\left(\mathrm{y}_i, \mathrm{y}_j\right) \rvert\}.
\label{eq:tgt_sentence_relevance}
\end{equation}

Finally, the ratio of target-side relevance to source-side relevance(TSSR) is calculated. A larger TSSR indicates a higher usage of target-side context in generating the next word $y_i$.

\begin{equation}
TSSR\left(\mathrm{y}_i\right)=\frac{R\left(\mathrm{y}_i\right)_{target-side}}{R\left(\mathrm{y}_i\right)_{source-side}}.
\label{eq:stab}
\end{equation}
Our final algorithm, referred to as Algorithm~\ref{alg:st_attention_bias}, is presented.
\begin{algorithm}[H]
\caption{Compute TSSR}
\label{alg:st_attention_bias}
\begin{algorithmic} 
\State \textbf{Input:} model, hypotheses sentence, source sentence, k
\State \textbf{Output:} TSSR
\For {$i$ in hypotheses sentence length}
    \If {$j<i$}
        \State Compute the relevance of next word $y_i$ and $y_j$ according to \ref{eq:tgt_token_relevance}
    \EndIf
\EndFor

\For {$i$ in source sentence length}
    \If {$j \leq i+k-1$}
        \State Compute the relevance of next word $y_i$ and $x_j$ according to \ref{eq:src_token_relevance}
    \EndIf
\EndFor

\State Compute Target-Side Relevance according to \ref{eq:tgt_sentence_relevance}

\State Compute Source-Side Relevance according to \ref{eq:src_sentence_relevance}

\State Compute TSSR according to \ref{eq:stab}

\end{algorithmic}
\end{algorithm}

TSSR is categorized into 10 intervals from 0 to INF, indicating the degree of Target Context Usage.

\subsection{The Relationship between Hallucination and Target-side Context Usage}

\begin{figure}[t]
  \centering
  \includegraphics[width=2.5in]{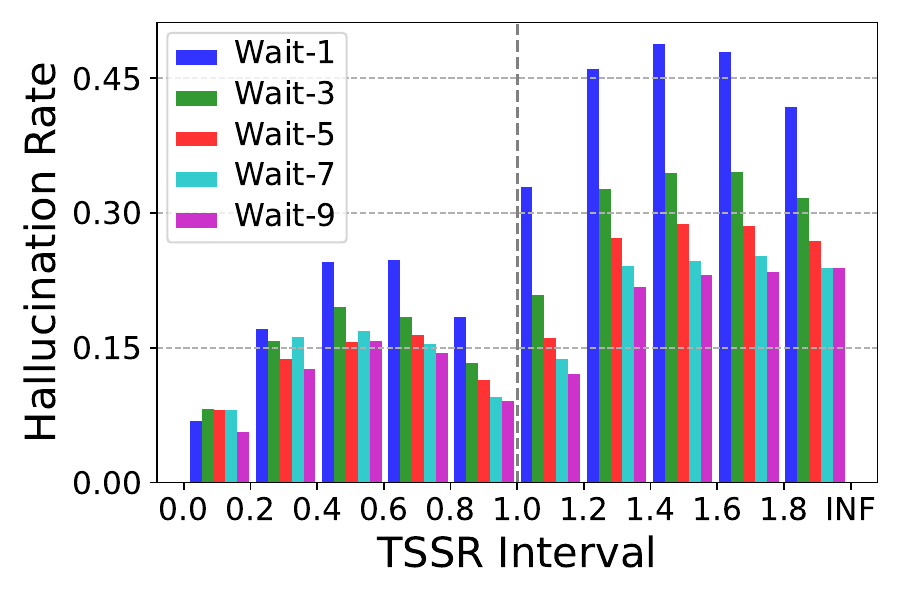}
  \captionof{figure}{HR on the valid set in different TSSR intervals of wait-$k$ models.}
  \label{fig2:DH_STAB}
\end{figure}
\paragraph{Using more target context leads to more severe hallucination.}
Initially, we analyze the relationship between a word's usage of the target-side context and its likelihood of being a hallucination. Building upon this, we explore the \textrm{HR} across different TSSR intervals, as depicted in Figure \ref{fig2:DH_STAB}.
Our findings demonstrate that in high TSSR intervals, \textrm{HR} is higher compared to low TSSR intervals. 
It indicates that a word using more target context is more likely to be a hallucination.

Further analysis revealed that when comparing different Wait-
values, there is a more pronounced increase in HR from low TSSR intervals to high TSSR intervals as k decreases, as depicted in Figure 2. This means that there maybe an increased likelihood of hallucinations occurring in words that are utilized with limited source-side context

\begin{figure}[t]
\centering
\subfigure[De-En]{
\includegraphics[width=2.0in]{
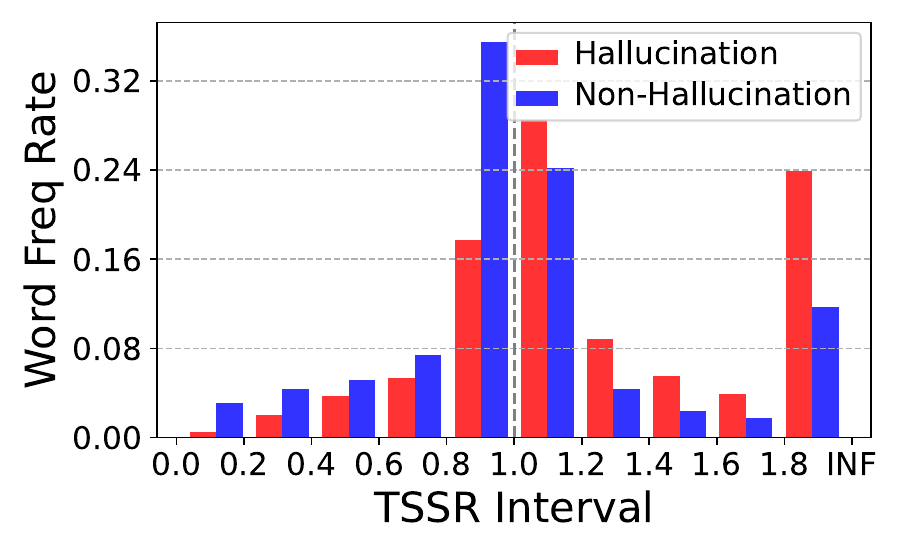
} 
\label{fig:hall_to_dependence_iwslt14deen}
}

\subfigure[Zh-En (human alignment annotation)]{
\includegraphics[width=2.0in]{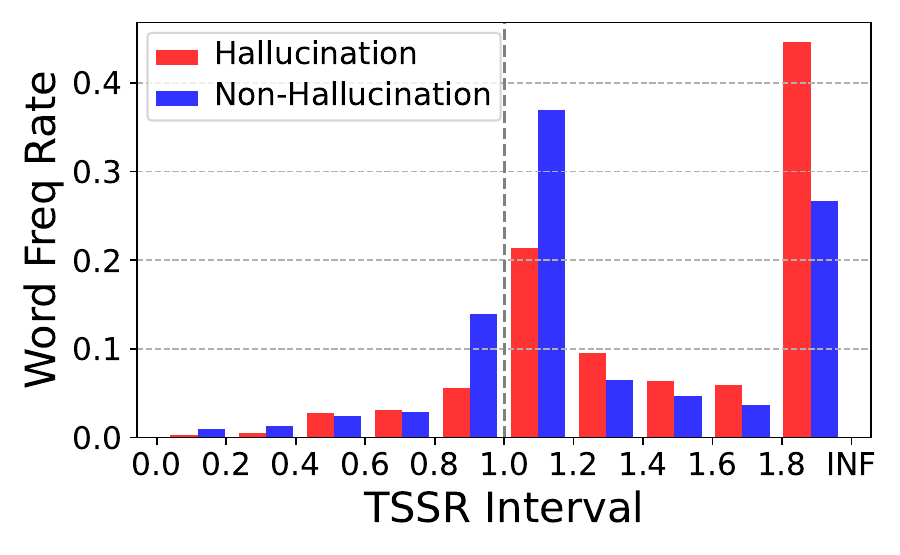} 
\label{fig:hall_to_dependence_mustczhen_annotation}
}
\caption{Word Frequency Rate of Hallucination and Non-Hallucination in different TSSR intervals for wait-$1$ model.}
\label{analysis.consistency_comparison}
\end{figure}

\paragraph{Hallucination words use more target context than Non-Hallucination words.}
The aforementioned analysis motivates us to investigate whether hallucination words indeed exhibit a higher usage of target-side context than non-hallucination words. To explore this, we analyze the TSSR distributions of hallucination and non-hallucination word frequencies.
Figure~\ref{fig:hall_to_dependence_iwslt14deen} reveals that hallucination words are concentrated on high TSSR intervals.
This means the model tends to use more target-side context for the generation of a hallucination word.
Furthermore, we observed that the word frequency rate of non-hallucination words is higher in the 0.8 \textasciitilde 1.2 TSSR range, also illustrated in Figure ~\ref{fig:hall_to_dependence_iwslt14deen}. Therefore, we propose that the model utilizes source-side context and target-side context similarly during the generation of non-hallucination words.
To further validate our claims of above analysis, we sample 100 sentences from the translation results of Zh-En using wait-$1$ decoding for human alignment annotation. We then conduct experiments similar to Figure~\ref{fig:hall_to_dependence_iwslt14deen}. The results as shown in Figure~\ref{fig:hall_to_dependence_mustczhen_annotation} are consistent with the conclusions drawn in automatic alignment annotation.


\subsection{Increasing Source-side Context Usage via Reducing Target-side Context Usage}

\begin{figure}[t]
  \centering
  \includegraphics[width=2.5in]{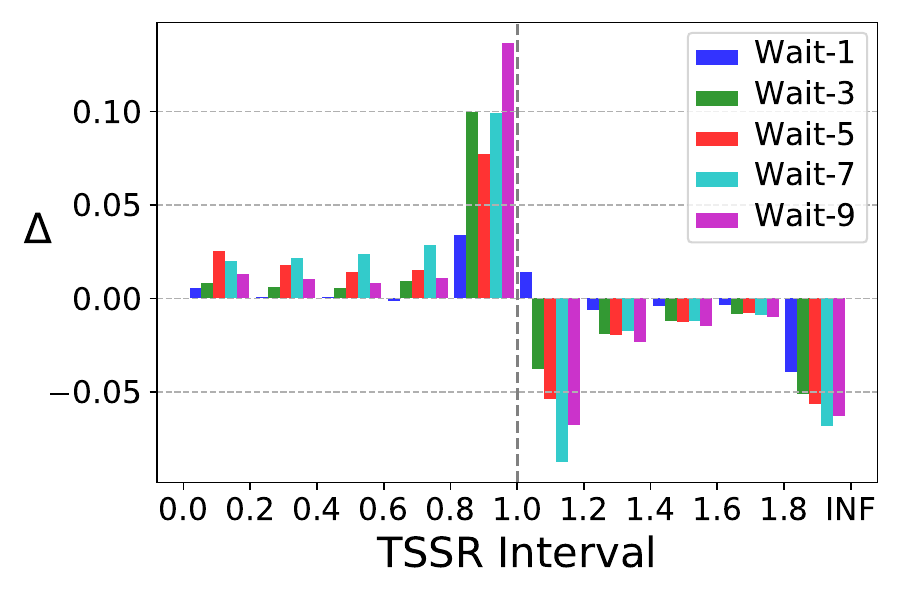}
  \captionof{figure}{Word Frequency Rate Change ($\Delta$) in different TSSR intervals with scheduled sampling training compared to the Baselines.}
  \label{fig:word_freq_rate_change_in_tssr1}
\end{figure}

\begin{table}[t]\centering 
  \setlength{\tabcolsep}{0.3em}
  \renewcommand{\arraystretch}{0.6}
  \resizebox{\linewidth}{!}{
  \begin{tabular}{lrrrrrr}
  \toprule
  \multicolumn{2}{l}{ } &  $k$=1  & $k$=3  & $k$=5 & $k$=7 & $k$=9  \\ 
  \midrule
  \multirow{2}{*}{Baselines} & BLEU $\uparrow$ &19.69 & 26.76 & 29.61 & 31.10 & 32.03  \\
     & \textrm{HR} $\% \downarrow$ &{31.28} & {22.57} & {18.58} & {16.41} & {15.21}  \\
  \midrule
  Scheduled-  & BLEU $\uparrow$ & 20.53 & 27.32 & 30.23 & 31.73 & 32.34 \\
  Sampling & \textrm{HR} $\% \downarrow$ & {30.85} & {21.62} & {17.84} & {15.16} & {13.84}  \\
  \bottomrule 
  \end{tabular}
  }
  \caption{BLEU scores and \textrm{HR} of wait-$k$ models.}
  \label{tab:bleu_and_hr_iwslt14deen}
\end{table}

\begin{figure}[t]
  \centering
  \includegraphics[width=2.5in]{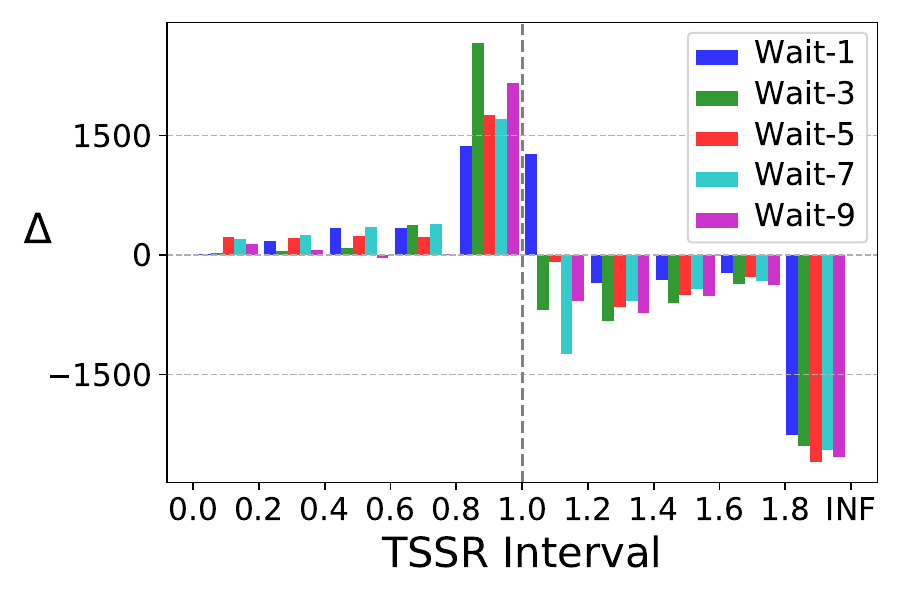}
  \captionof{figure}{Hallucination Frequency Change ($\Delta$) in different TSSR intervals with scheduled sampling training compared to the Baselines.}
  \label{fig:hall_freq_change_in_tssr1}
\end{figure}

Observing the association between hallucination and usage of target-side context, we posit that reducing this reliance might be a viable approach to mitigate the hallucination in SiMT.
Inspired by \cite{bengio2015scheduled,zhang-etal-2019-bridging}, we adopt the scheduled sampling training to guide the models to pay more attention on the source-side context by adding noise to the target-side context. 
Specifically, we randomly replace the ground truth tokens with predicted ones using a decaying probability. 
The results shown in Figure~\ref{fig:word_freq_rate_change_in_tssr1} indicate a decrease in target-context usage and an increase in source-context usage.
Scheduled sampling training exhibits improvements in BLEU scores and reductions in \textrm{HR} as presented in Table~\ref{tab:bleu_and_hr_iwslt14deen}.
It successfully reduces hallucination words using more target-side context, but also indirectly increases hallucination words using more source-side context, as shown in Figure~\ref{fig:hall_freq_change_in_tssr1}. 
Therefore, a better method to flexibly handle the usage between target-side and source-side context is required.

\section{Related Work}
In NMT, previous works have delved into the phenomenon of hallucinations\cite{lee_hallucinations,muller2020domain,wang2020exposure,raunak2021curious,zhou2021detecting}.
Specifically, \citet{voita_analyzing_2021} assessed the relative contributions of source and target context to predictions.
\citet{weng2020towards,miao_prevent_2021} argued that an important reason for hallucination is the model's excessive attention to partial translations in NMT. 
Furthermore, \citet{guerreiro2022looking} conducted a comprehensive study of hallucinations in NMT. Differing from these works focusing on NMT, this paper conducts a comprehensive analysis of hallucination in SiMT.

\section{Conclusions}
This paper conducts the first comprehensive analysis of hallucinations in SiMT from two perspectives: understanding the hallucination words from both frequency and predictive distributions and their effects on the usage of target-context information. 
Intensive Experiments demonstrate some valuable findings: 1) the frequency distribution of hallucination words is with high entropy and their predictive distribution is with high uncertainty due to the difficulty in memorizing hallucination words during training. 2) hallucination words make use of more target-side context than source-side context, and it is possible to alleviate hallucination by decreasing the usage of target-side context.

\section*{Limitations}
We highlight four main limitations of our work.

Firstly, instead of focusing on more recent adaptive policy, our analysis focuses on the hallucinations in the Wait-$k$ Policy~\citep{ma2018stacl}, which is the most widely used fixed policy in SiMT to ensure a simple and familiar setup that is easy to reproduce and generalize.

Secondly, although we propose a simple methods to control the usage of target information, attempting to mitigate the hallucination in SiMT, we only achieve limited improvement. 
In the future, we will explore more flexible and robust approaches for controlling target context usage to better mitigate the hallucination and achieve greater performance.

A further limitation of our study is that we exclusively analyze hallucinations as defined in Section 2, without considering detached hallucinations. This omission arises from the absence of established and reliable automated evaluation methods for detecting such detached hallucinated words.

Moreover, our study is constrained by its reliance on aligner tools, potentially introducing alignment biases. Therefore, when applying our approach to datasets with lower alignment accuracy, careful consideration is warranted regarding the necessity for additional validation and adjustment.

\section*{Acknowledgements}
We would like to thank the anonymous reviewers and meta-reviewer for their insightful suggestions.
The work was supported by the National Natural Science Foundation of China under Grant 62276077, Guangdong Basic and Applied Basic Research Foundation (2024A1515011205), and Shenzhen College Stability Support Plan under Grants GXWD20220811170358002 and GXWD20220817123150002.

\bibliography{ARR/new_simt}

\appendix

\appendix
\section{Detailed Experimental Settings}
\label{detailed exp}
On IWSLT'14 De$\leftrightarrow$En, we train on 160K pairs, develop on 7K held out pairs.
All data is tokenized and lower-cased and we segment sequences using byte pair encoding \cite{Sennrich16acl} with 10K merge operations. The resulting vocabularies are of 8.8K and 6.6K types in German and English respectively. 


On MuST-C Release V2.0 Zh$\rightarrow$En\footnote{\url{https://ict.fbk.eu/must-c-release-v2-0/}}, we train on 358,853 pairs, develop on 1,349 pairs. 
Jieba\footnote{\url{https://github.com/fxsjy/jieba}} are employed for Chinese word segmentation. All data is tokenized by SentencePiece resulting in 32k word vocabularies in Chinese and English.

Following~\citet{elbayad2020efficient} and~\citet{zhang2021universal}, We train Transformer Small on IWSLT14 De$\rightarrow$En.
We train Transformer Base on MuST-C Release V2.0 Zh$\rightarrow$En.

\section{Experimental Results on IWSLT14 En $\rightarrow$ De Dataset}

\subsection{Results of Word Frequency Distribution on IWSLT14 De$\rightarrow$En Dataset}
\label{Distribution}

\begin{figure}[H]
  \centering
  \includegraphics[width=2.0in]{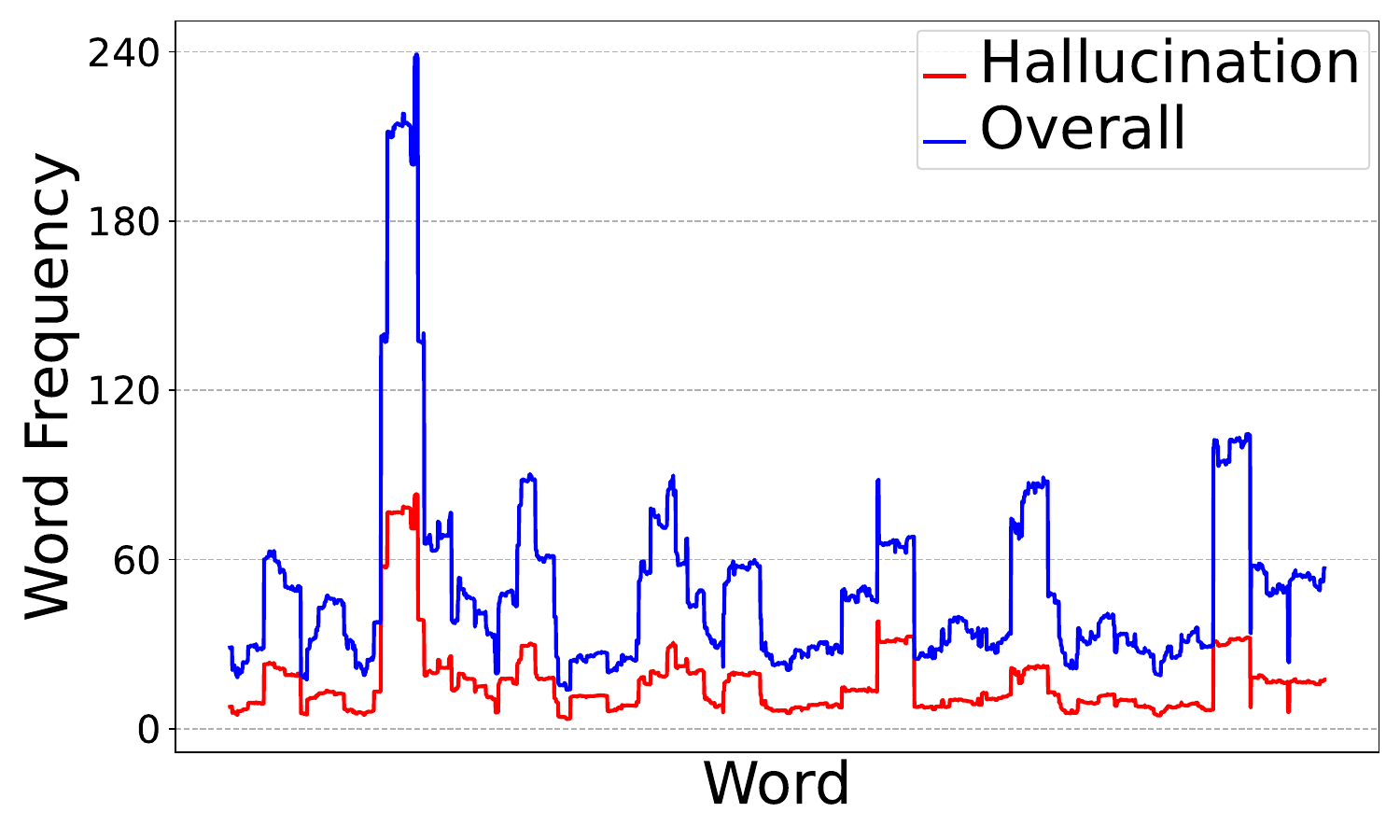}
  \captionof{figure}{Word frequency of Hallucination and Overall on IWSLT14 De$\rightarrow$En valid hypotheses set of wait-$1$.}
  \label{fig:word_freq_hall_vs_overall_wait1app}
\end{figure}

\begin{figure}[H]
  \centering
  \includegraphics[width=2.0in]{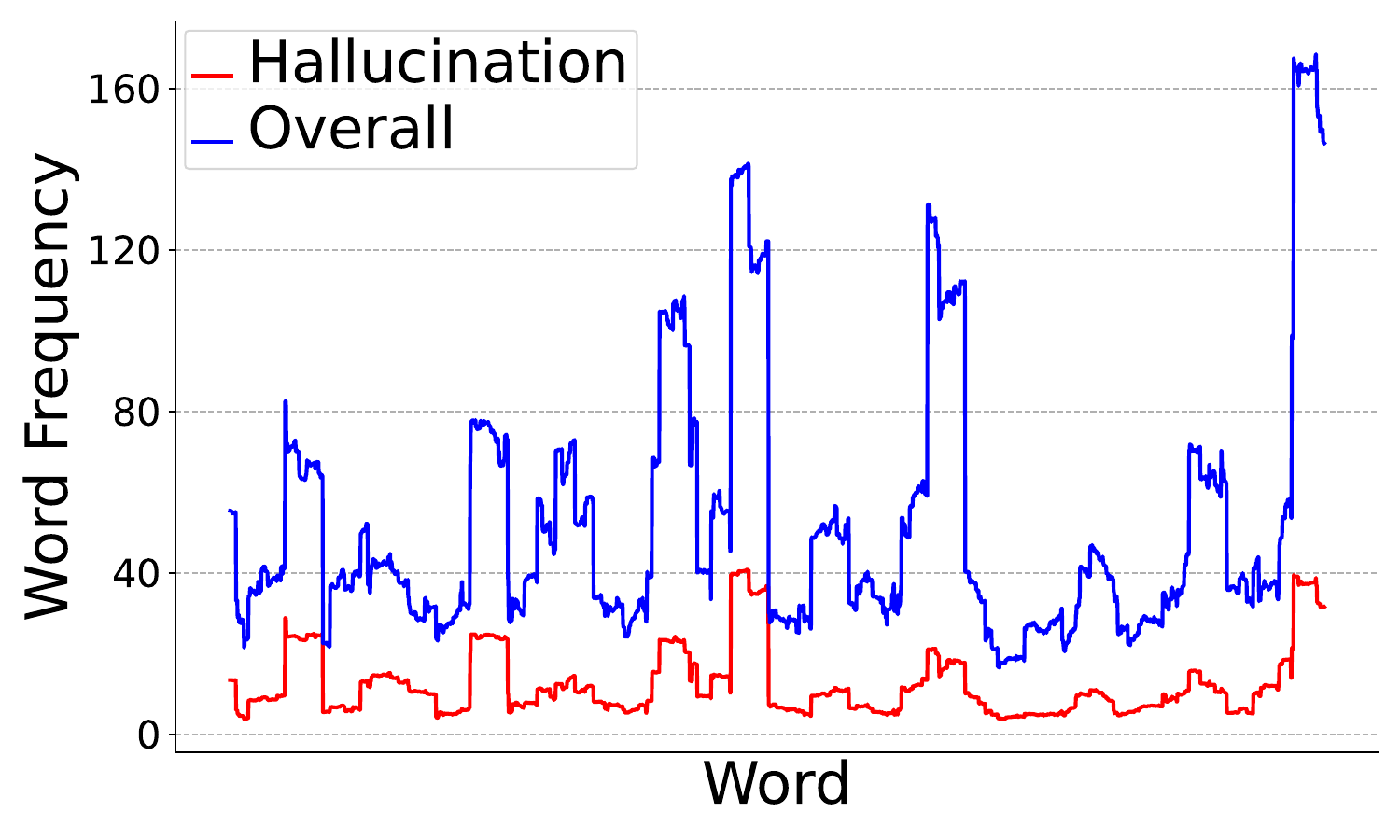}
  \captionof{figure}{Word frequency of Hallucination and Overall on IWSLT14 De$\rightarrow$En valid hypotheses set of wait-$3$.}
  \label{fig:word_freq_hall_vs_overall_wait3}
\end{figure}

\begin{figure}[H]
  \centering
  \includegraphics[width=2.0in]{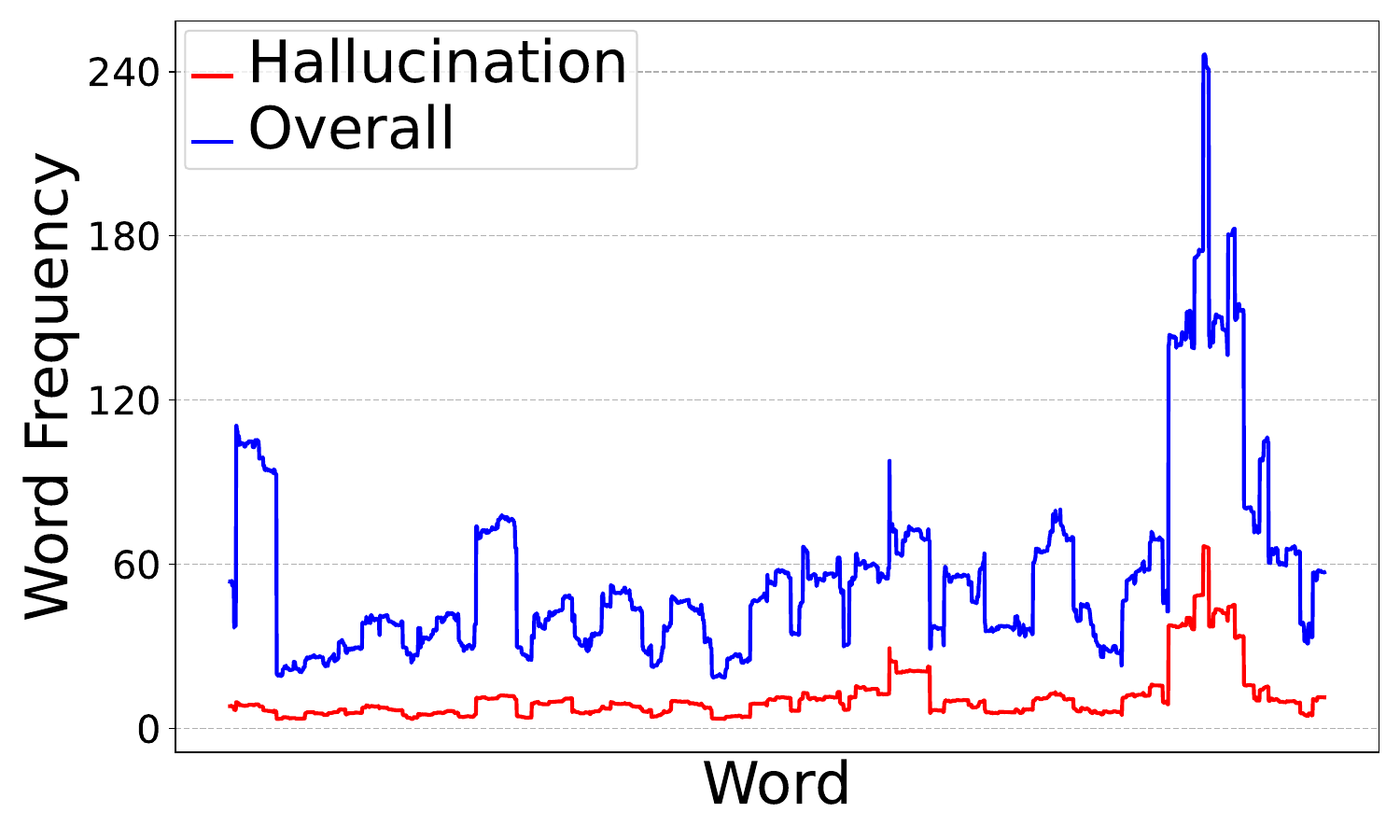}
  \captionof{figure}{Word frequency of Hallucination and Overall on IWSLT14 De$\rightarrow$En valid hypotheses set of wait-$5$.}
  \label{fig:word_freq_hall_vs_overall_wait5}
\end{figure}

\begin{figure}[H]
  \centering
  \includegraphics[width=2.0in]{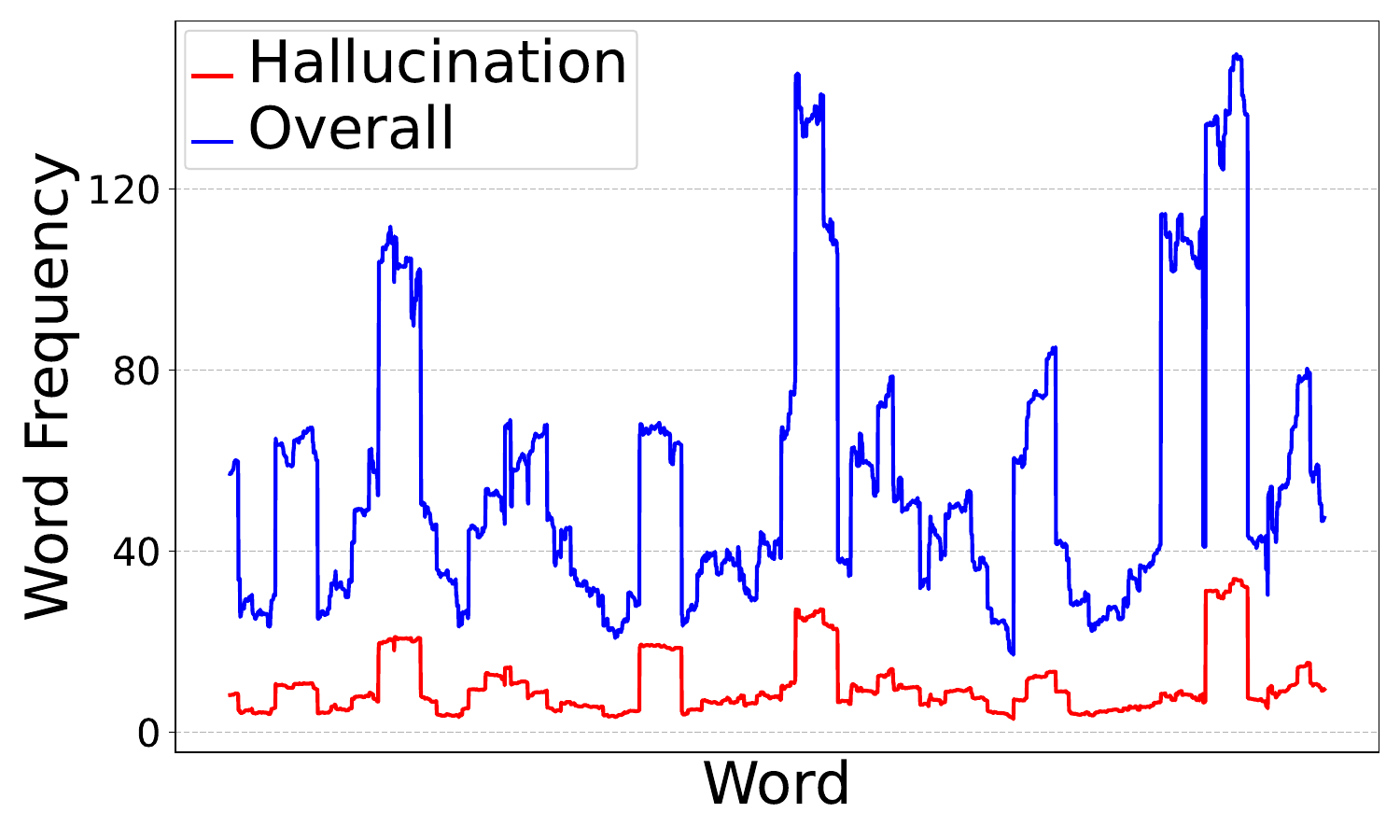}
  \captionof{figure}{Word frequency of Hallucination and Overall on IWSLT14 De$\rightarrow$En valid hypotheses set of wait-$7$.}
  \label{fig:word_freq_hall_vs_overall_wait7}
\end{figure}

\begin{figure}[H]
  \centering
  \includegraphics[width=2.0in]{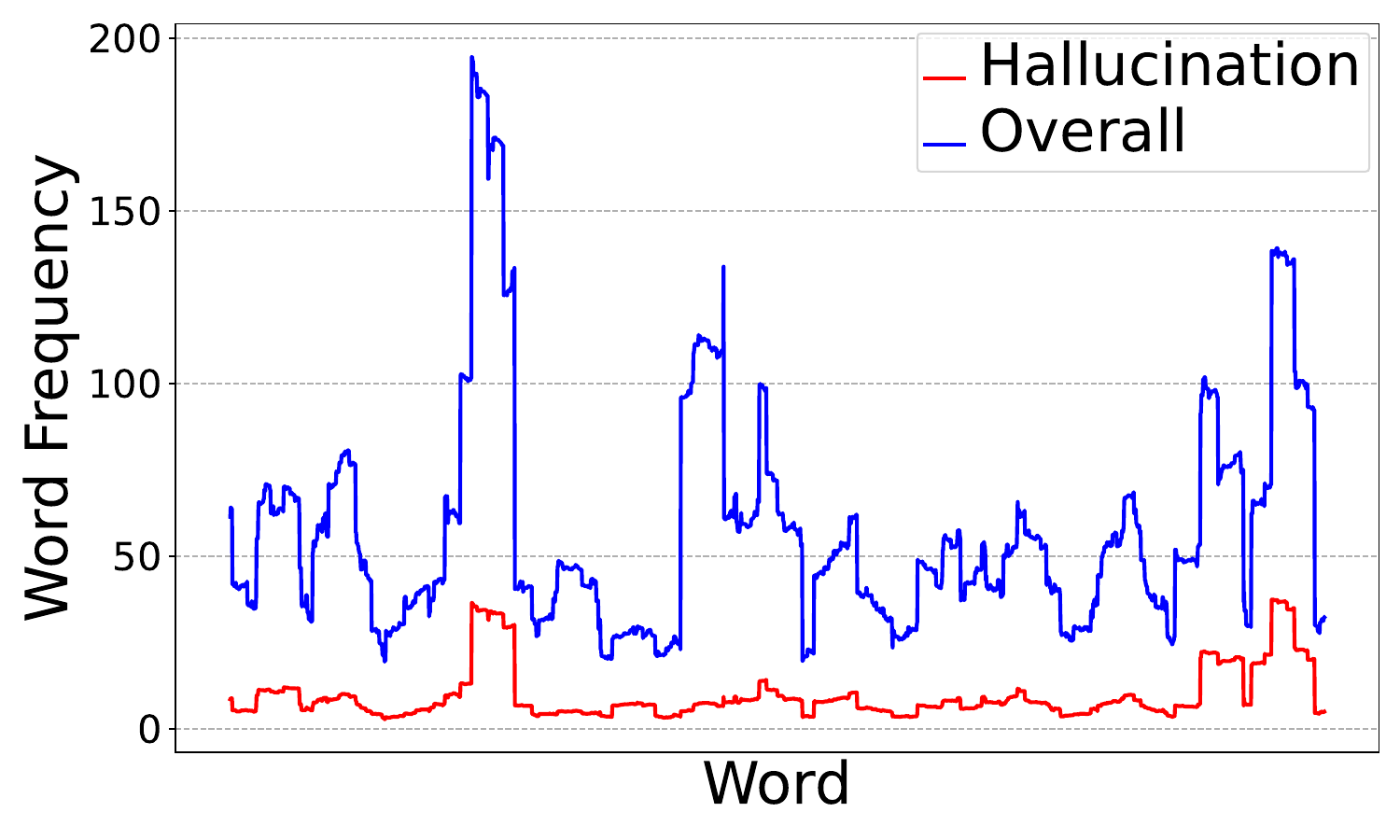}
  \captionof{figure}{Word frequency of Hallucination and Overall on IWSLT14 De$\rightarrow$En valid hypotheses set of wait-$9$.}
  \label{fig:word_freq_hall_vs_overall_wait9}
\end{figure}


\subsection{Results of Word Frequency Rate in TSSR on IWSLT14 De$\rightarrow$En Dataset}
\label{word_freq}

\begin{figure}[H]
  \centering
  \includegraphics[width=2.0in]{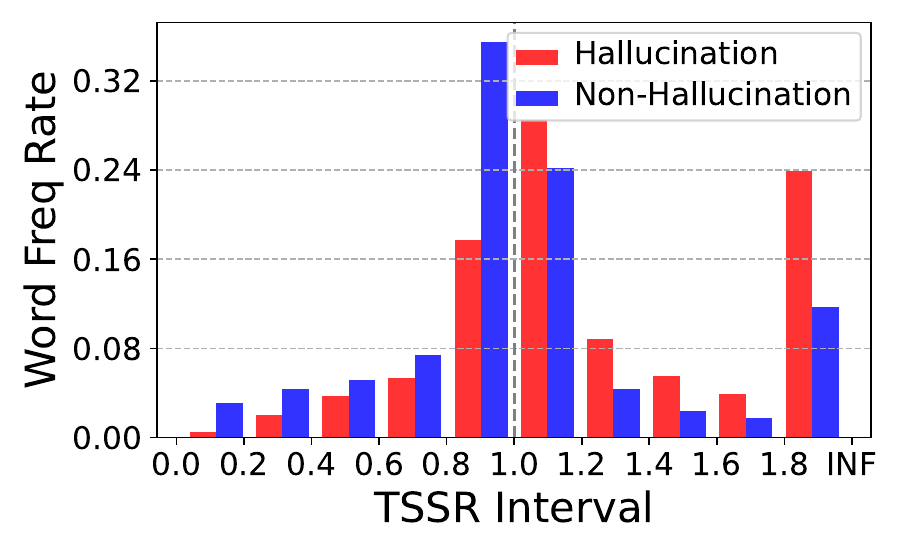}
  \captionof{figure}{Word Frequency Rate of Hallucination and Non-Hallucination in different TSSR intervals for the wait-$1$ model.}
  \label{fig:hall_to_dependence.1}
\end{figure}

\begin{figure}[H]
  \centering
  \includegraphics[width=2.0in]{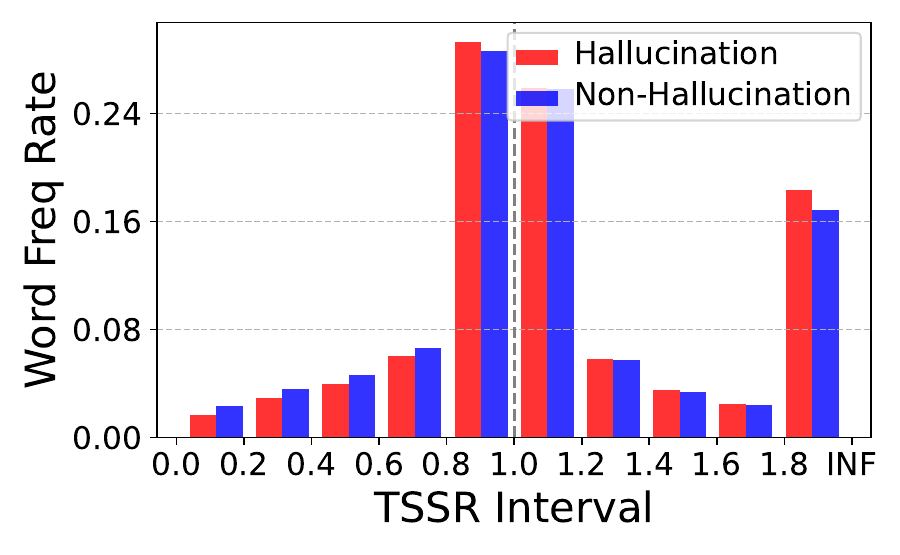}
  \captionof{figure}{Word Frequency Rate of Hallucination and Non-Hallucination in different TSSR intervals for the wait-$1$ model with WSPAlign Annotation \cite{wu2023wspalign}.}
  \label{fig:hall_to_dependence.1.wsp}
\end{figure}

\begin{figure}[H]
  \centering
  \includegraphics[width=2.0in]{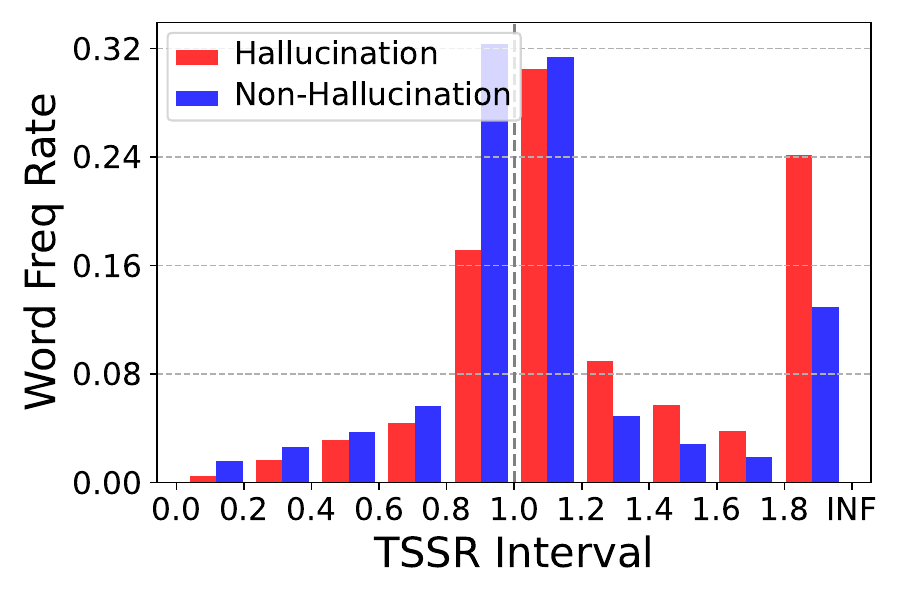}
  \captionof{figure}{Word Frequency Rate of Hallucination and Non-Hallucination in different TSSR intervals for the wait-$3$ model.}
  \label{fig:hall_to_dependence.3}
\end{figure}

\begin{figure}[H]
  \centering
  \includegraphics[width=2.0in]{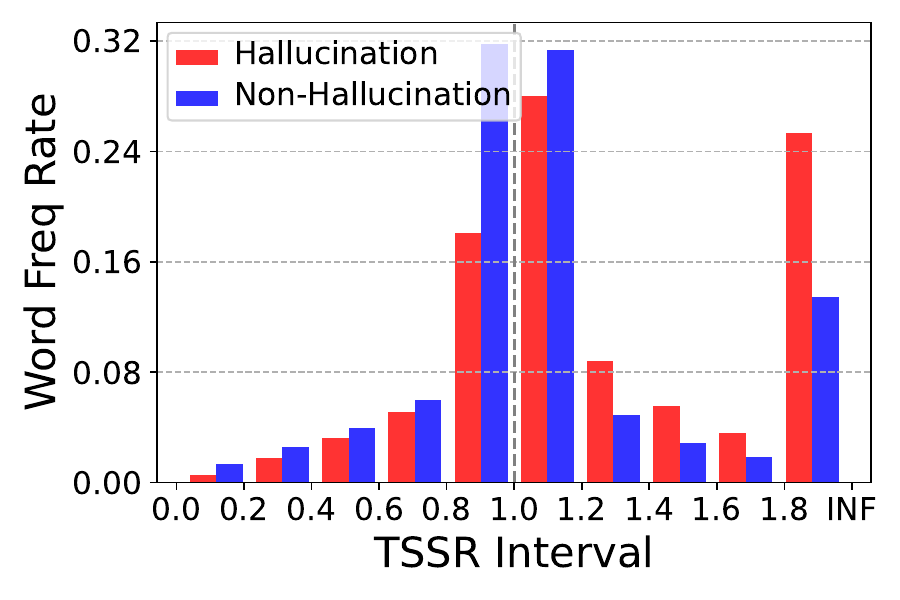}
  \captionof{figure}{Word Frequency Rate of Hallucination and Non-Hallucination in different TSSR intervals for the wait-$5$ model.}
\end{figure}

\begin{figure}[H]
  \centering
  \includegraphics[width=2.0in]{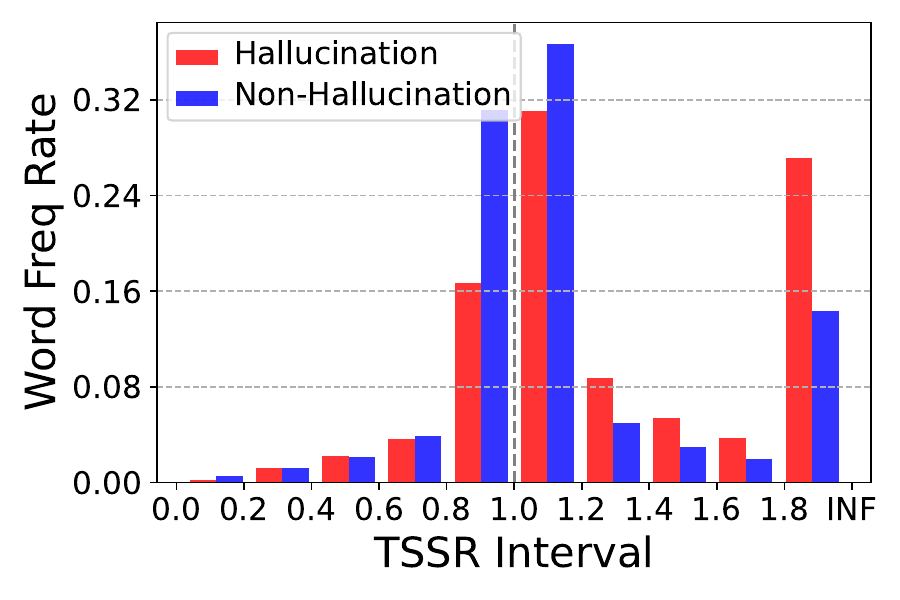}
  \captionof{figure}{Word Frequency Rate of Hallucination and Non-Hallucination in different TSSR intervals for the wait-$7$ model.}
\end{figure}

\begin{figure}[H]
  \centering
  \includegraphics[width=2.0in]{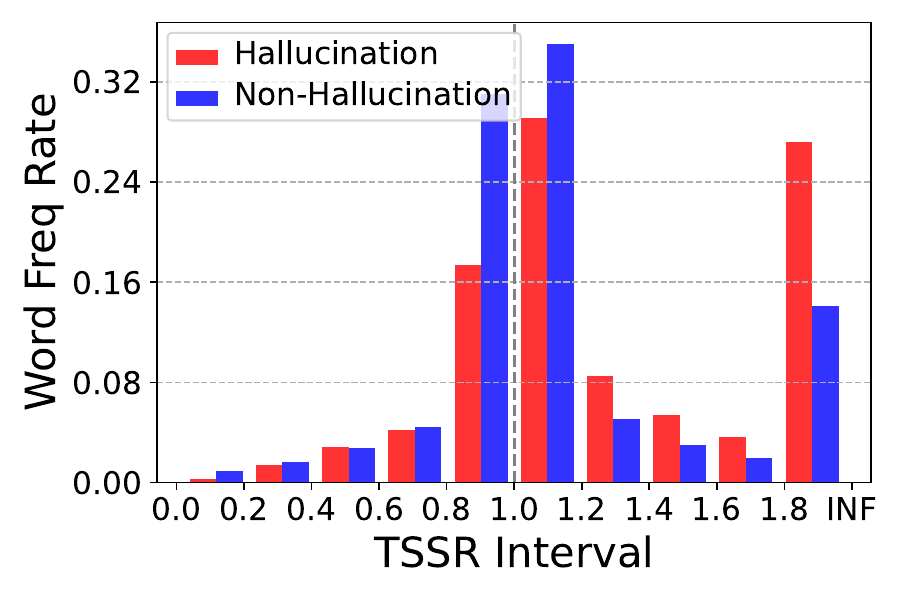}
  \captionof{figure}{Word Frequency Rate of Hallucination and Non-Hallucination in different TSSR intervals for wait-$9$ model.}
\end{figure}

\section{Experimental Results on IWSLT14 En$\rightarrow$De Dataset}
\label{iwslt14ende}
\begin{figure}[H]
  \centering
  \includegraphics[width=2.0in]{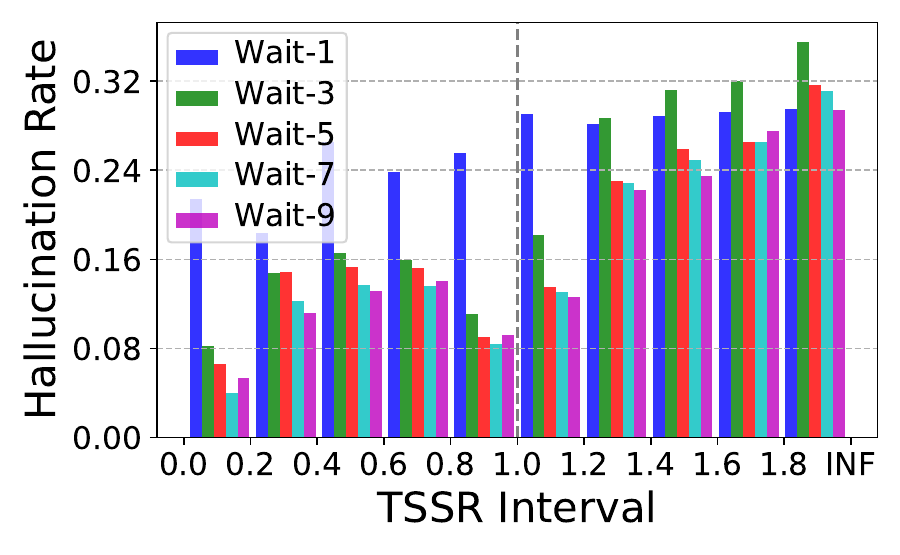}
  \captionof{figure}{HR on the valid set in different TSSR intervals of wait-$k$ models.}
\end{figure}

\begin{figure}[H]
  \centering
  \includegraphics[width=2.0in]{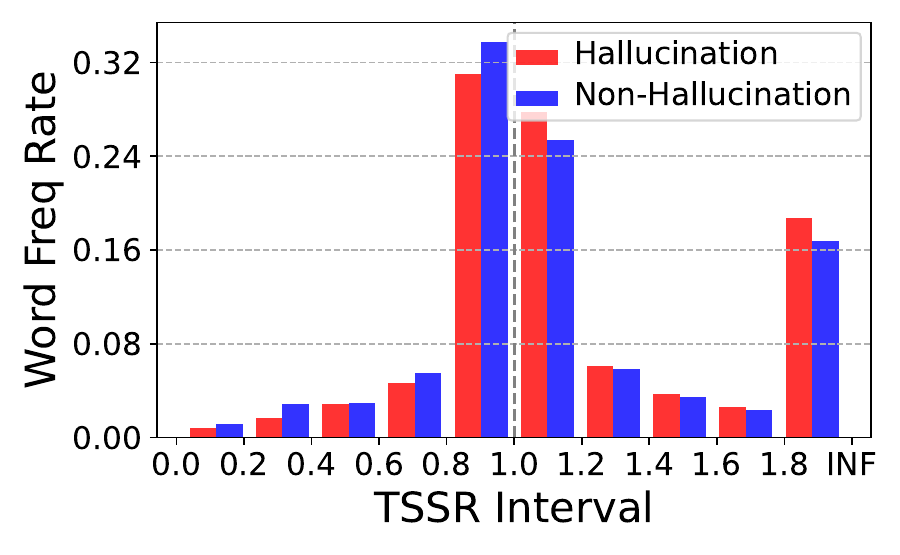}
  \captionof{figure}{Word Frequency Rate of Hallucination and Non-Hallucination in different TSSR intervals for the wait-$1$ model.}
\end{figure}

\begin{figure}[H]
  \centering
  \includegraphics[width=2.0in]{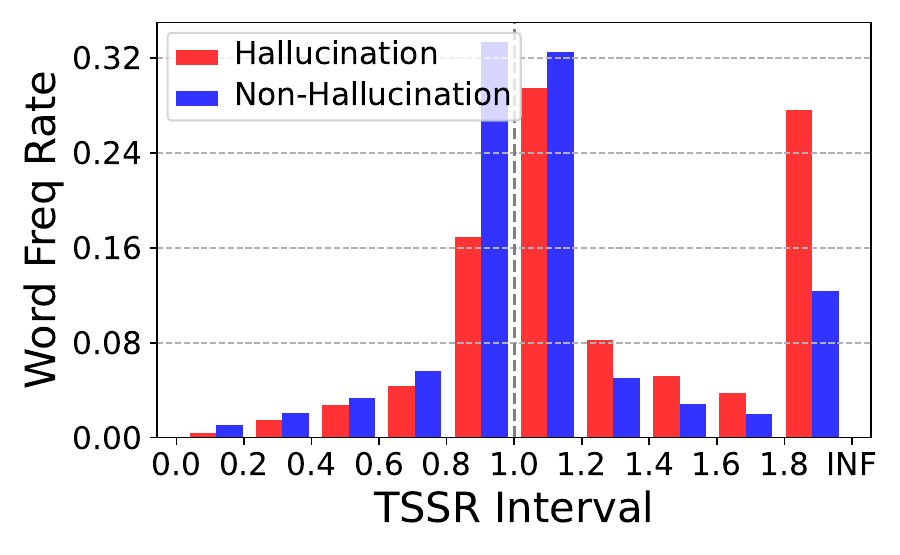}
  \captionof{figure}{Word Frequency Rate of Hallucination and Non-Hallucination in different TSSR intervals for the wait-$3$ model.}
\end{figure}

\begin{figure}[H]
  \centering
  \includegraphics[width=2.0in]{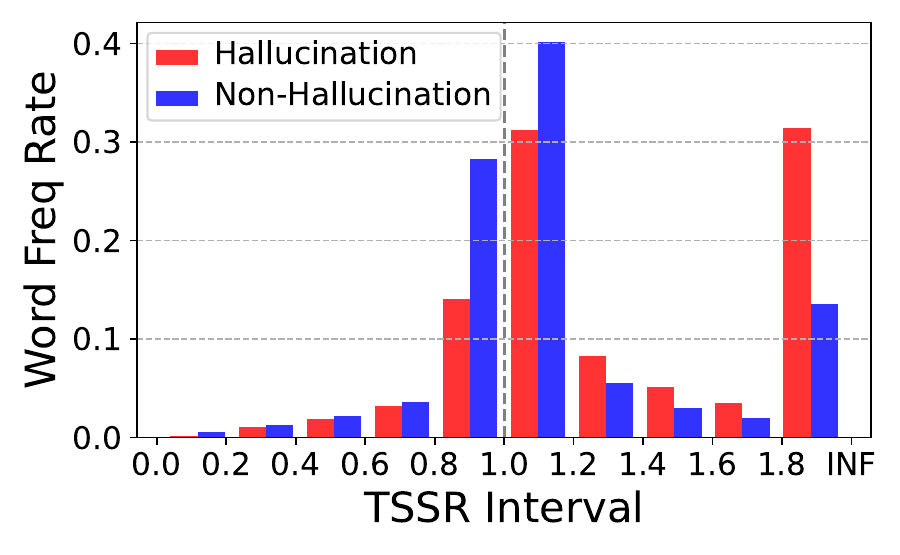}
  \captionof{figure}{Word Frequency Rate of Hallucination and Non-Hallucination in different TSSR intervals for the wait-$5$ model.}
\end{figure}

\begin{figure}[H]
  \centering
  \includegraphics[width=2.0in]{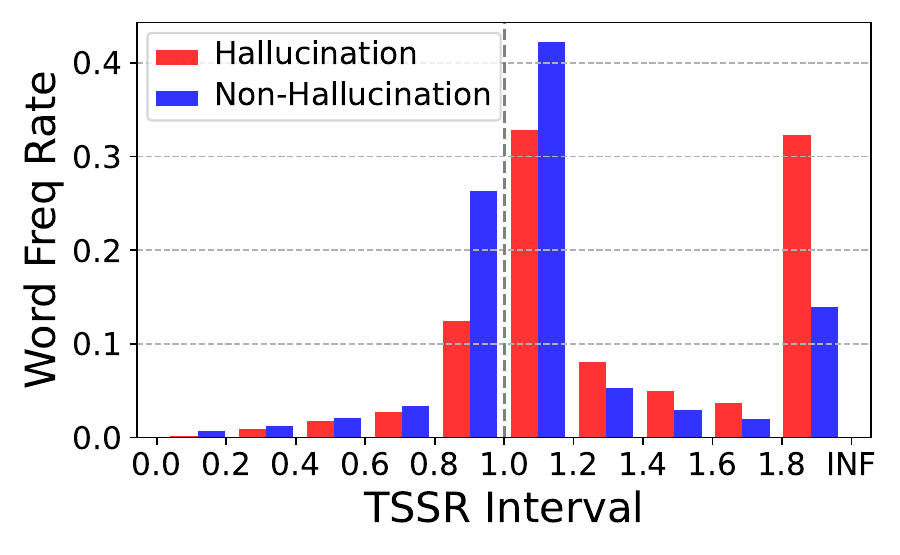}
  \captionof{figure}{Word Frequency Rate of Hallucination and Non-Hallucination in different TSSR intervals for the wait-$7$ model.}
\end{figure}

\begin{figure}[H]
  \centering
  \includegraphics[width=2.0in]{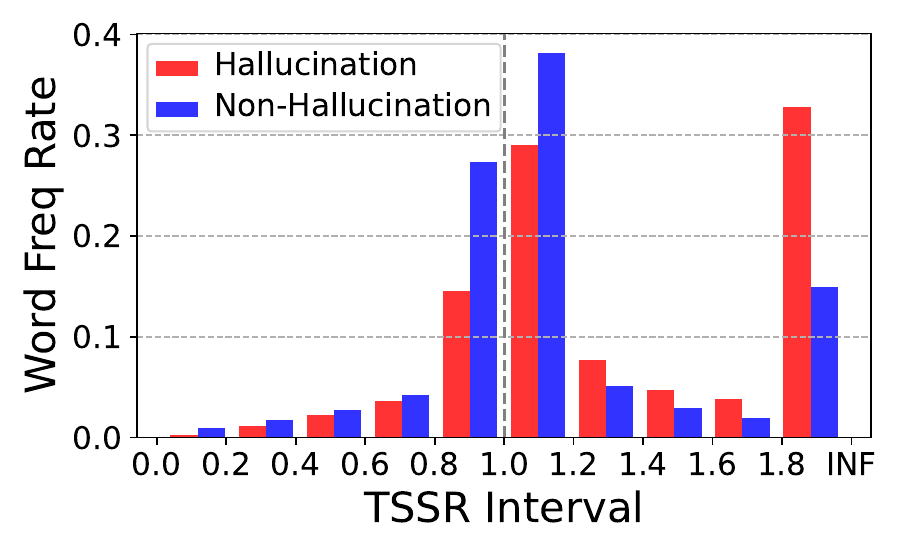}
  \captionof{figure}{Word Frequency Rate of Hallucination and Non-Hallucination in different TSSR intervals for wait-$9$ model.}
\end{figure}

\begin{table}[H]\centering 
  \setlength{\tabcolsep}{0.3em}
  \renewcommand{\arraystretch}{0.6}
  \resizebox{\linewidth}{!}{
  \begin{tabular}{lrrrrrr}
  \toprule
  \multicolumn{2}{l}{ } &  $k$=1  & $k$=3  & $k$=5 & $k$=7 & $k$=9  \\ 
  \midrule
  \multirow{2}{*}{Baselines} & BLEU $\uparrow$ &15.75 & 22.03 & 24.99 & 26.22 & 26.60  \\
     & \textrm{HR} $\% \downarrow$ &{27.46} & {19.73} & {16.72} & {16.24} & {15.93}  \\
  \midrule
  Scheduled-  & BLEU $\uparrow$ & 16.83 & 22.78 & 25.80 & 26.98 & 27.41 \\
  Sampling & \textrm{HR} $\% \downarrow$ & {26.19} & {18.58} & {15.66} & {14.96} & {14.81}  \\
  \bottomrule 
  \end{tabular}
  }
  \caption{BLEU scores and \textrm{HR} of wait-$k$ models.}
  \label{tab:bleu_and_hr}
\end{table}

\begin{figure}[H]
  \centering
  \includegraphics[width=2.0in]{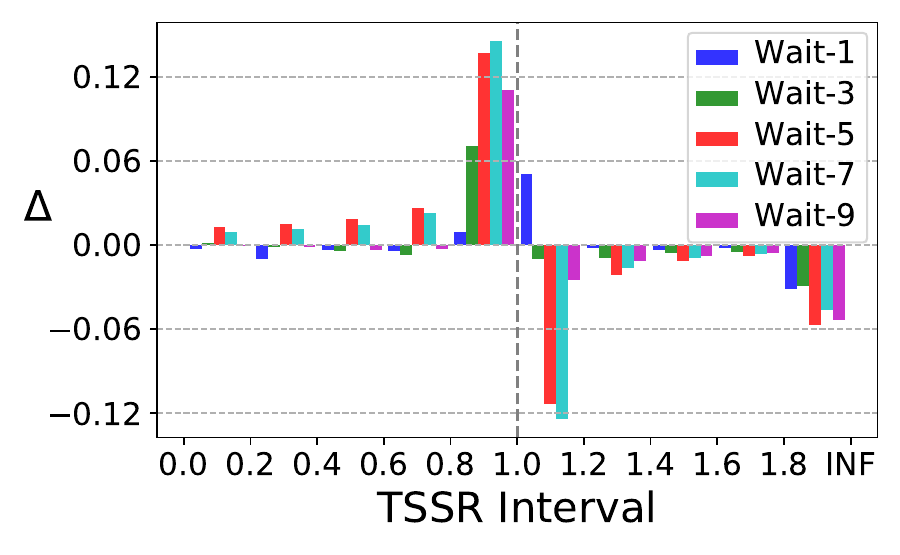}
  \captionof{figure}{Word Frequency Rate Change ($\Delta$) in different TSSR intervals with scheduled sampling training compared to the Baselines.}
  \label{fig:word_freq_rate_change_in_tssr_ende}
\end{figure}

\begin{figure}[H]
  \centering
  \includegraphics[width=2.0in]{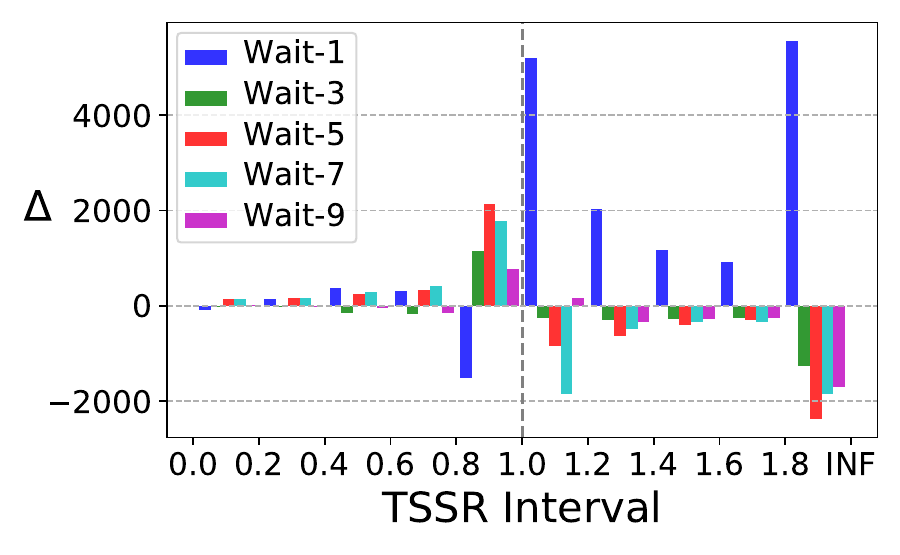}
  \captionof{figure}{Hallucination Frequency Change ($\Delta$) in different TSSR intervals with scheduled sampling training compared to the Baselines.}
  \label{fig:hall_freq_change_in_tssr}
\end{figure}

\section{Experimental Results on MuST-C Zh$\rightarrow$En Dataset}
\label{mustczhen}

\begin{table}[H]
  \centering
  \begin{adjustbox}{width=\columnwidth}
  \begin{tabular}{c|cccccc}
    \toprule
    \textbf{$k$} & $1$ & $3$ & $5$ & $7$ & $9$ & $\infty$ \\
    \midrule
    \textrm{HR} $\%$ &{33.96} & {25.31} & {23.22} & {21.84} & {20.73} & 19.43 \\
    \bottomrule
  \end{tabular}
  \end{adjustbox}
  \caption{\textrm{HR} on MuST-C Zh$\rightarrow$En valid set of wait-$k$, where $k=\infty$ means Full-sentence MT.}
\end{table}

\begin{figure}[H]
  \centering
  \includegraphics[width=2.0in]{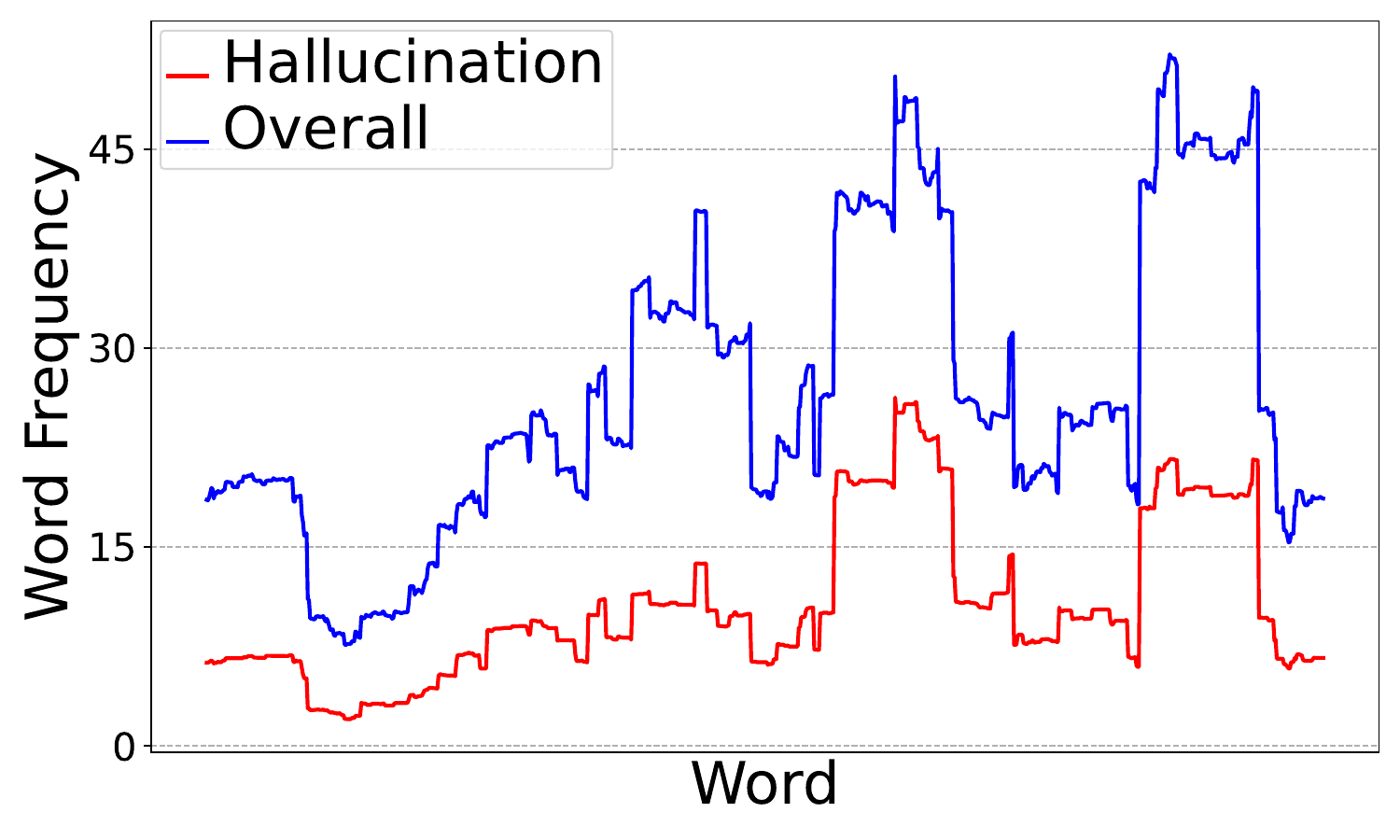}
  \captionof{figure}{Word frequency of Hallucination and Overall on valid hypotheses set of wait-$1$.}
\end{figure}

\begin{figure}[H]
  \centering
  \includegraphics[width=2.0in]{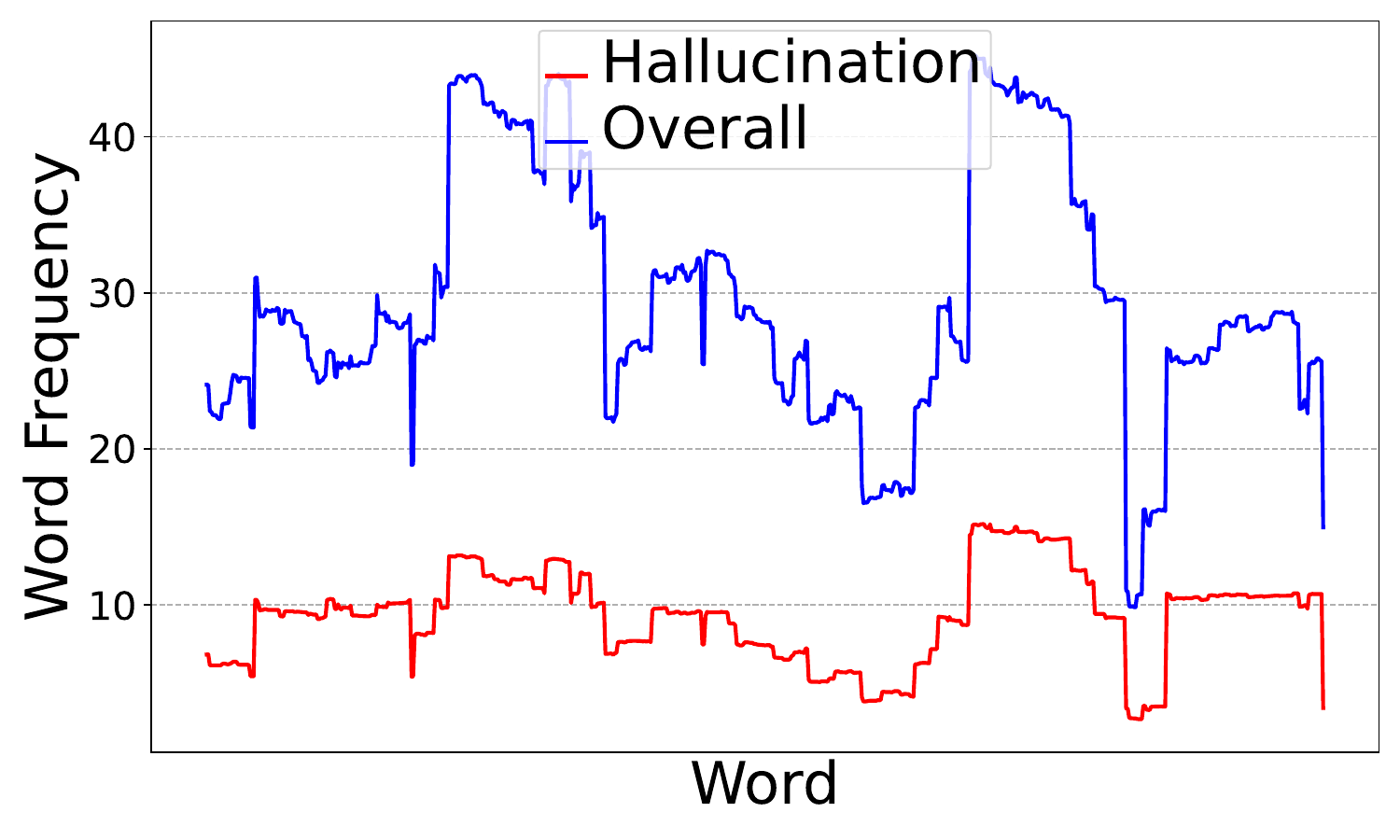}
  \captionof{figure}{Word frequency of Hallucination and Overall on valid hypotheses set of wait-$3$.}
\end{figure}

\begin{figure}[H]
  \centering
  \includegraphics[width=2.0in]{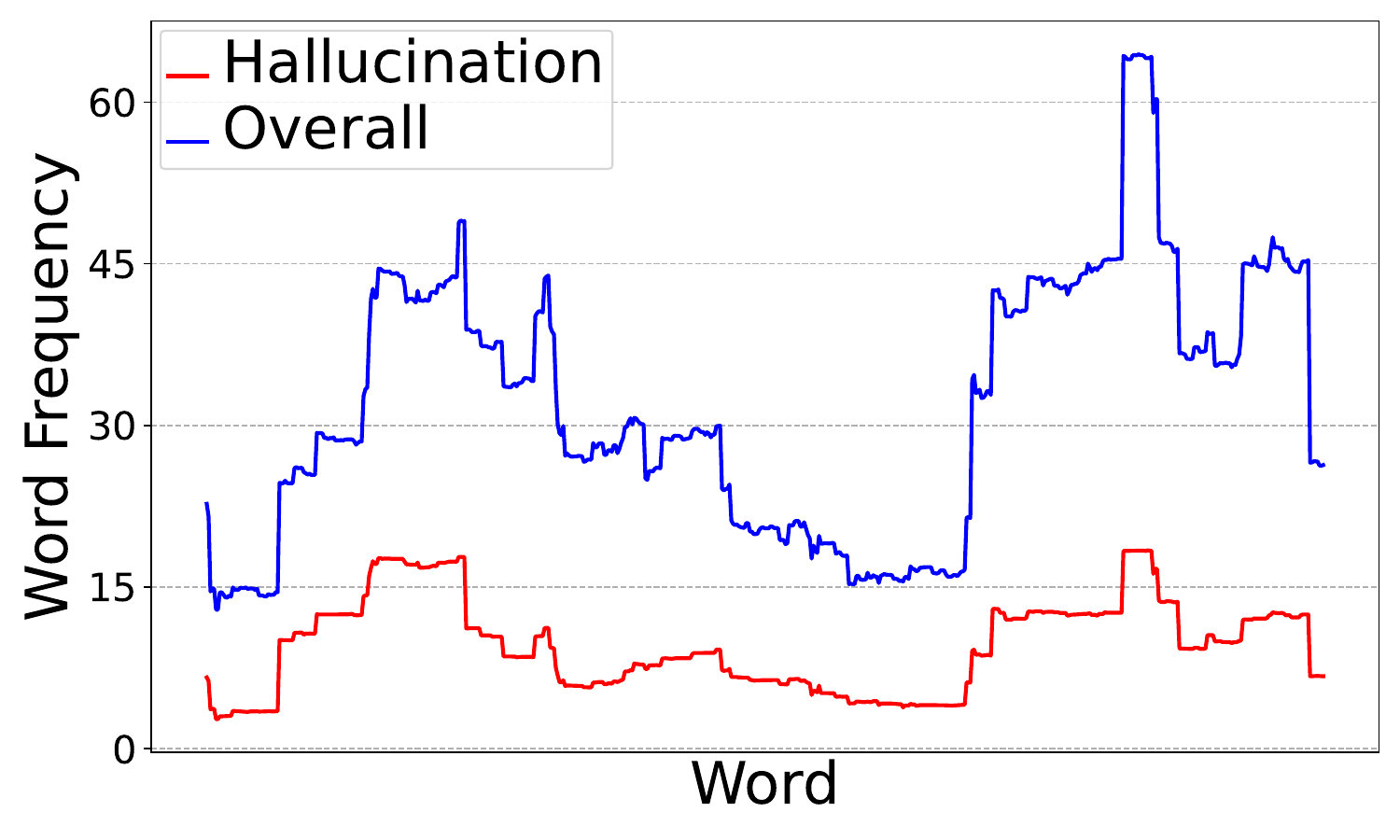}
  \captionof{figure}{Word frequency of Hallucination and Overall on valid hypotheses set of wait-$5$.}
\end{figure}

\begin{figure}[H]
  \centering
  \includegraphics[width=2.0in]{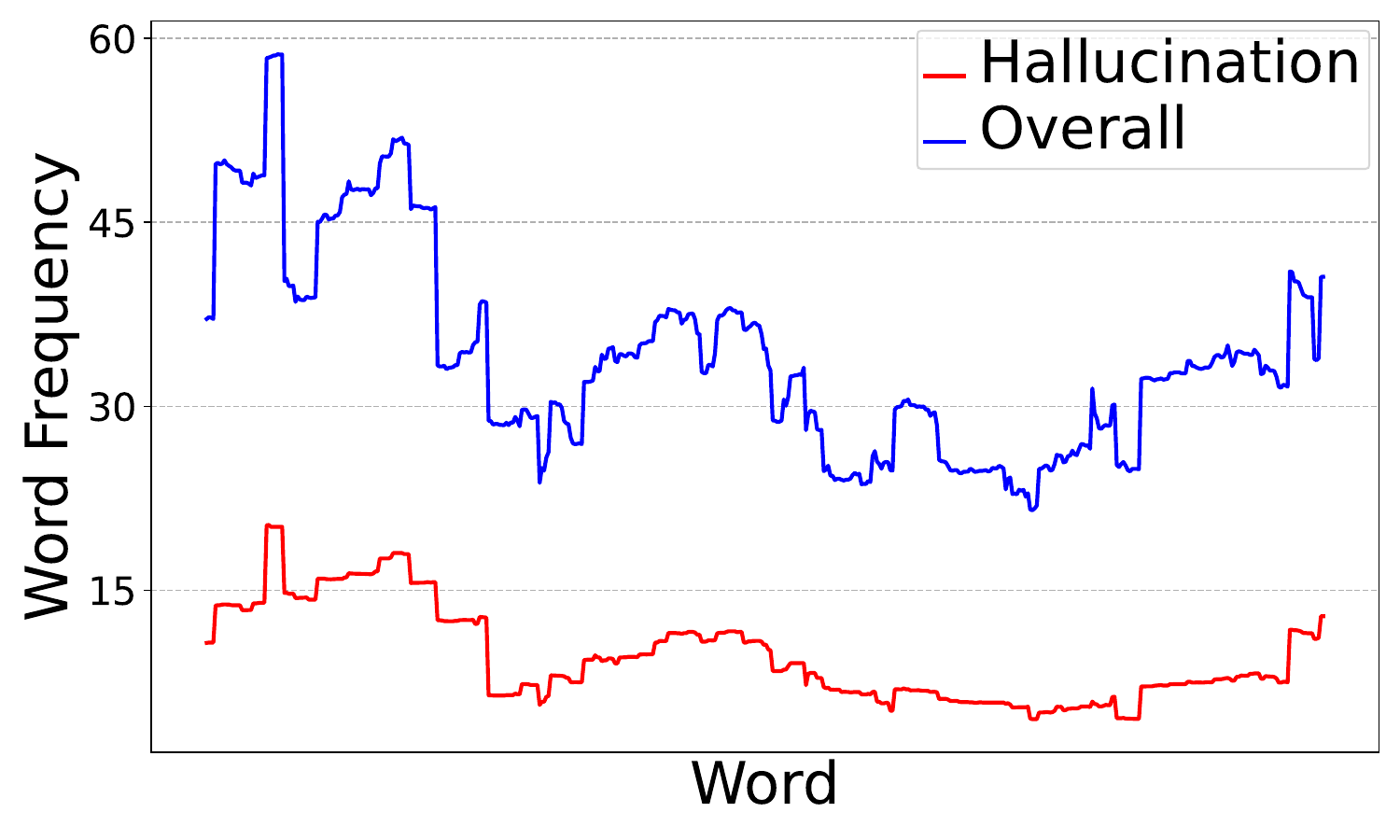}
  \captionof{figure}{Word frequency of Hallucination and Overall on valid hypotheses set of wait-$7$.}
\end{figure}

\begin{figure}[H]
  \centering
  \includegraphics[width=2.0in]{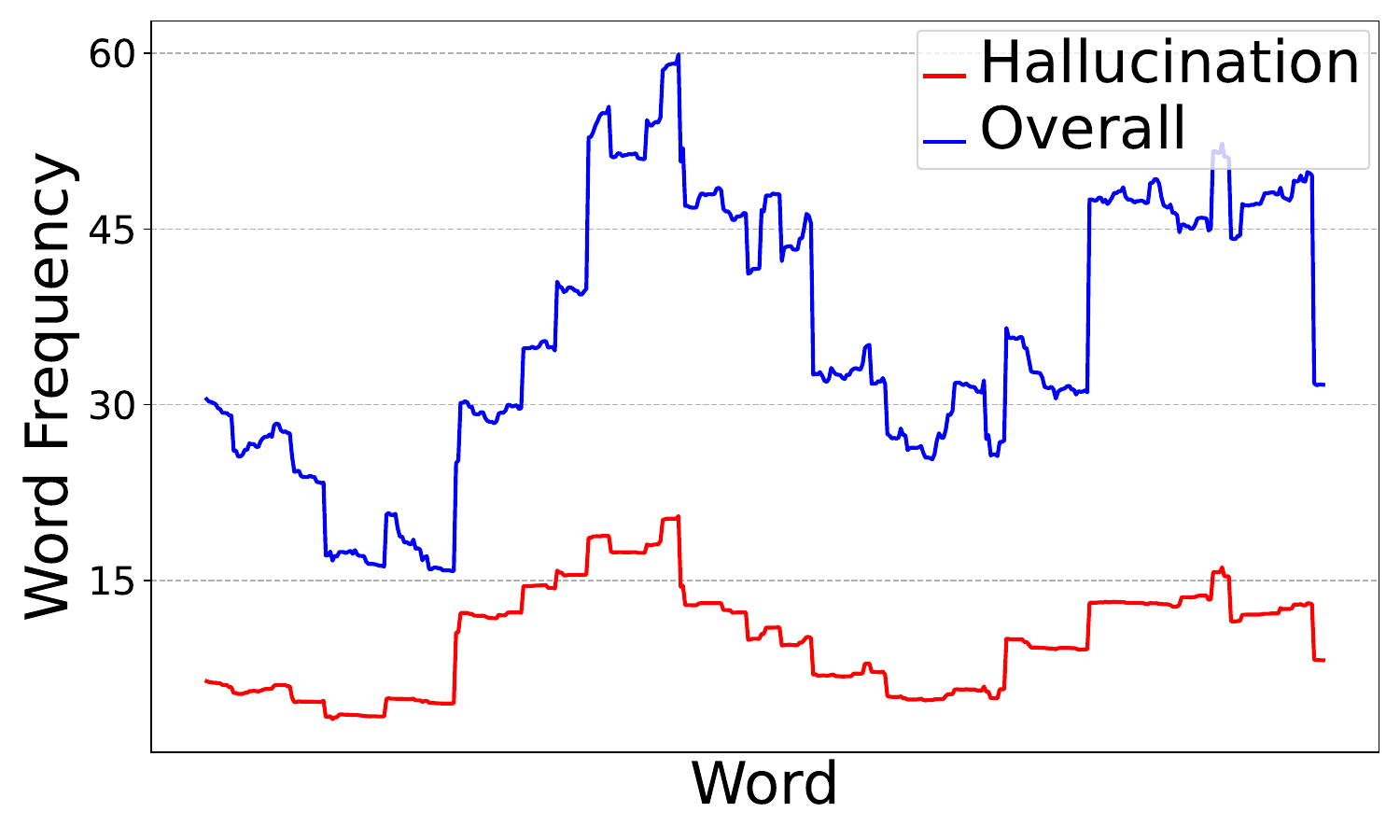}
  \captionof{figure}{Word frequency of Hallucination and Overall on valid hypotheses set of wait-$9$.}
\end{figure}

\begin{table}[H]
  \centering
  \begin{adjustbox}{width=\columnwidth}
  \begin{tabular}{c|ccccc}
    \toprule
    \textbf{$k$} & $1$ & $3$ & $5$ & $7$ & $9$  \\
    \midrule
    Hallucination & 6.57 & 6.52 & 6.35 & 6.29 & 6.23 \\
    Overall & 8.23 & 8.44 & 8.49 & 8.53 & 8.52 \\
    \bottomrule
  \end{tabular}
  \end{adjustbox}
  \caption{Word frequency distribution entropy of Hallucination and Overall on MuST-C Zh$\rightarrow$En valid hypotheses set of wait-$k$.}
\end{table}

\begin{table}[H]
  \centering
  \begin{adjustbox}{width=0.85\columnwidth}
    \begin{tabular}{l *{3}{S[table-format=2, table-space-text-post=\%]}}
      \toprule
      & {Train Ref} & {Valid Ref} & {Valid Hypo} \\
      \midrule
      Train Ref    & 1.00          & 0.25         & 0.18                 \\
      Valid Ref    & 0.25         & 1.00          & 0.54                 \\
      Valid Hypo & 0.18         & 0.54         & 1.00                  \\
      \bottomrule
    \end{tabular}
  \end{adjustbox}
  \caption{The correlation between the \textrm{HR} of words on the Valid Hypotheses (Valid Hypo), Valid Reference (Valid Ref) and Train Reference (Train Ref) of $H_{wait-1}(t,a)$.}
  \label{tab:correlation_matrix}
\end{table}

\begin{table}[H]
\centering
\resizebox{1.0\columnwidth}{!}{
\begin{tabular}{lcccccccc}
\toprule
\multirow{4}{*}{\bf Wait-$k$}                    & \multicolumn{4}{c}{\bf Valid set} & \multicolumn{4}{c}{\bf Training subset} \\ 
\cmidrule(lr){2-5}\cmidrule(lr){6-9}
&  \multicolumn{2}{c}{\bf Uncertainty}        &  \multicolumn{2}{c}{\bf Confidence}       &  \multicolumn{2}{c}{\bf Uncertainty}        &  \multicolumn{2}{c}{\bf Confidence}       \\ 
\cmidrule(lr){2-3}\cmidrule(lr){4-5}\cmidrule(lr){6-7}\cmidrule(lr){8-9}
&\bf  H        &\bf  NH       &\bf  H        &\bf  NH   & \bf  H        &\bf  NH       &\bf  H        &\bf  NH       \\ 
\midrule
\multicolumn{1}{c}{$k$=1}     & 3.23         & 2.70       & 0.44         & 0.54     & 3.27                & 2.34 & 0.44         & 0.60   \\ 
\multicolumn{1}{c}{$k$=3}     & 3.00         & 2.43       & 0.49         & 0.58     & 2.91                & 2.14 & 0.50         & 0.63   \\ 
\multicolumn{1}{c}{$k$=5}     & 2.67         & 2.33       & 0.53         & 0.60     & 2.59                & 2.00 & 0.55         & 0.65   \\ 
\multicolumn{1}{c}{$k$=7}     & 2.64         & 2.32       & 0.54         & 0.60    & 2.50                & 2.00 & 0.56         & 0.65    \\ 
\multicolumn{1}{c}{$k$=9}     & 2.60         & 2.29       & 0.55         & 0.60     & 2.44                & 2.00 & 0.57         & 0.65   \\ 
\bottomrule
\end{tabular}}
\caption{The Uncertainty and Confidence of Hallucination ({\bf H}) and Non-Hallucination ({\bf NH}) on the valid set and training subset of wait-$k$ models.}
\end{table}

\begin{figure}[H]
  \centering
  \includegraphics[width=2.0in]{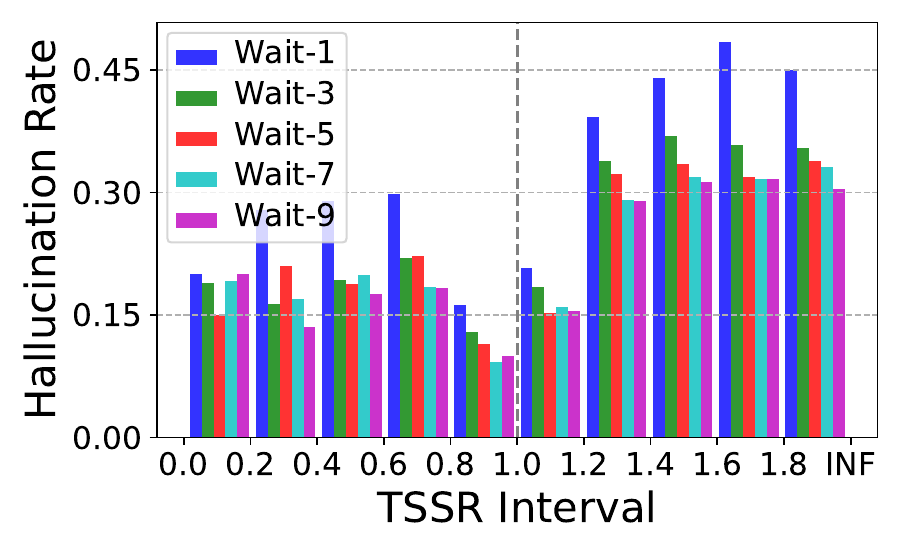}
  \captionof{figure}{HR on the valid set in different TSSR intervals of wait-$k$ models.}
\end{figure}

\begin{figure}[H]
  \centering
  \includegraphics[width=2.0in]{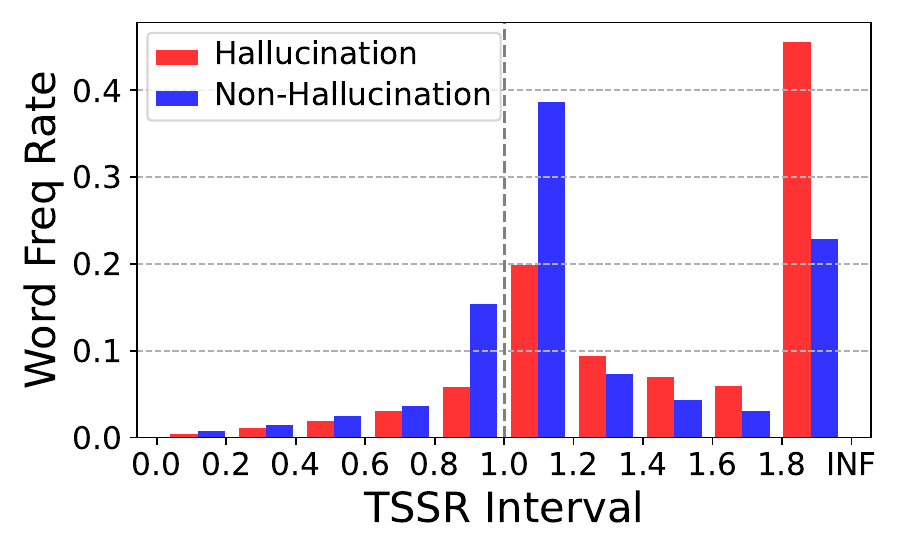}
  \captionof{figure}{Word Frequency Rate of Hallucination and Non-Hallucination in different TSSR intervals for the wait-$1$ model.}
\end{figure}

\begin{figure}[H]
  \centering
  \includegraphics[width=2.0in]{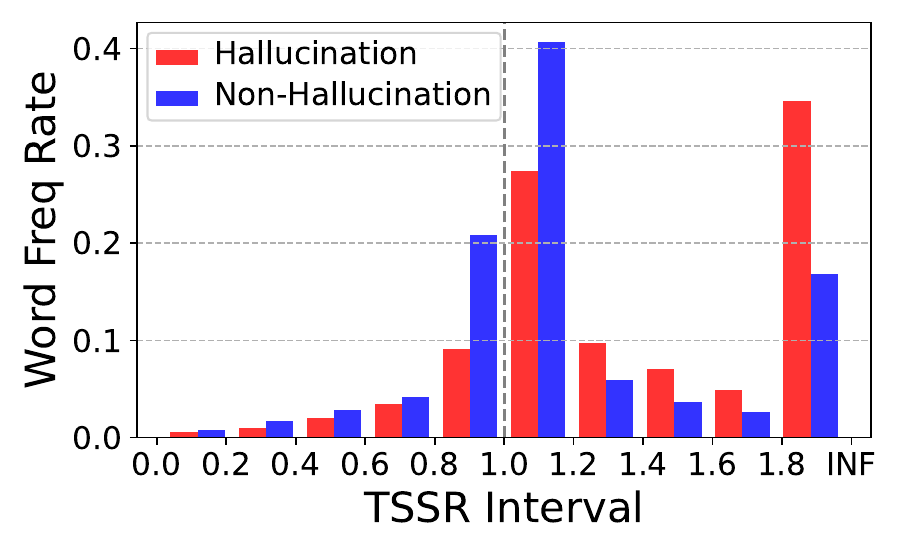}
  \captionof{figure}{Word Frequency Rate of Hallucination and Non-Hallucination in different TSSR intervals for the wait-$3$ model.}
\end{figure}

\begin{figure}[H]
  \centering
  \includegraphics[width=2.0in]{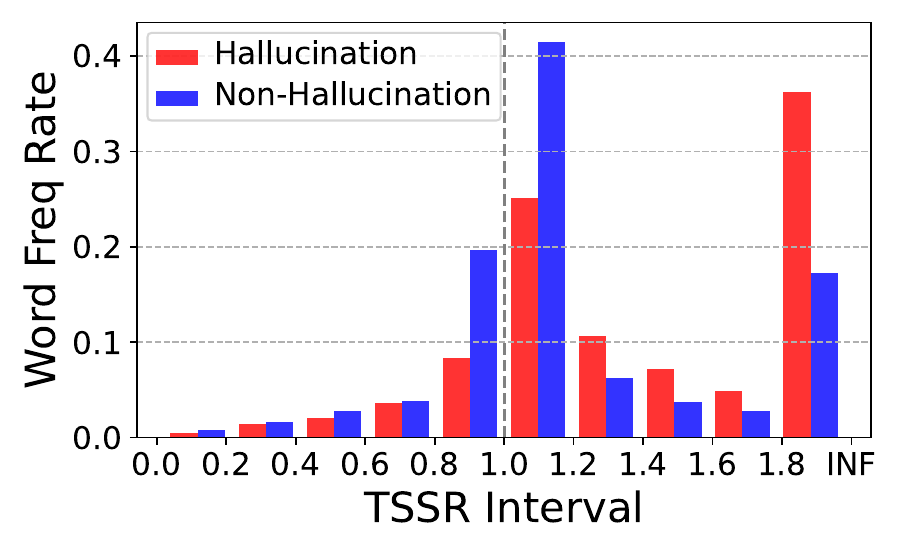}
  \captionof{figure}{Word Frequency Rate of Hallucination and Non-Hallucination in different TSSR intervals for the wait-$5$ model.}
\end{figure}

\begin{figure}[H]
  \centering
  \includegraphics[width=2.0in]{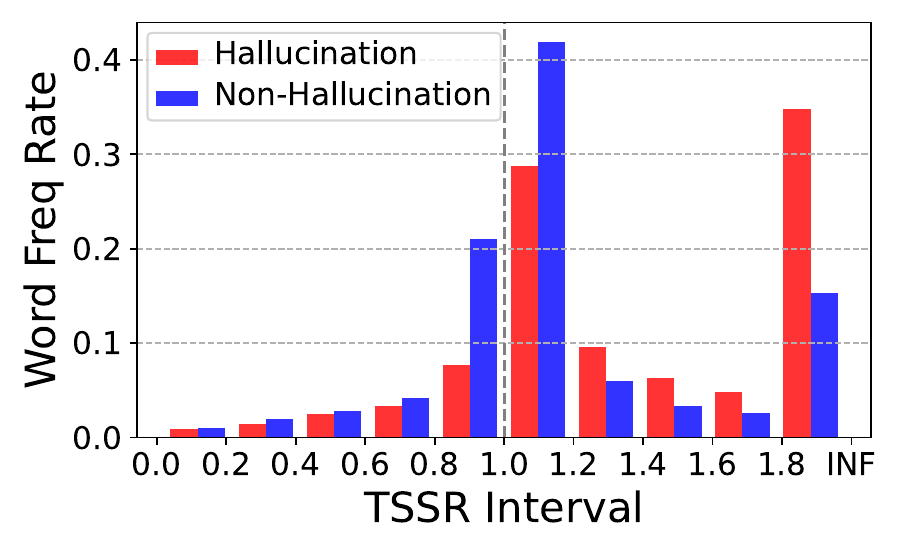}
  \captionof{figure}{Word Frequency Rate of Hallucination and Non-Hallucination in different TSSR intervals for the wait-$7$ model.}
\end{figure}

\begin{figure}[H]
  \centering
  \includegraphics[width=2.0in]{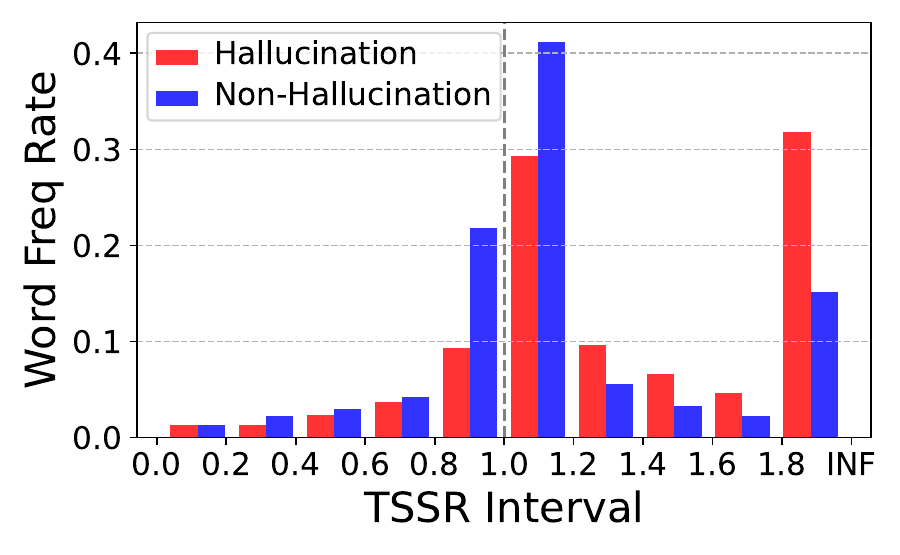}
  \captionof{figure}{Word Frequency Rate of Hallucination and Non-Hallucination in different TSSR intervals for wait-$9$ model.}
\end{figure}

\begin{figure}[H]
  \centering
  \includegraphics[width=2.0in]{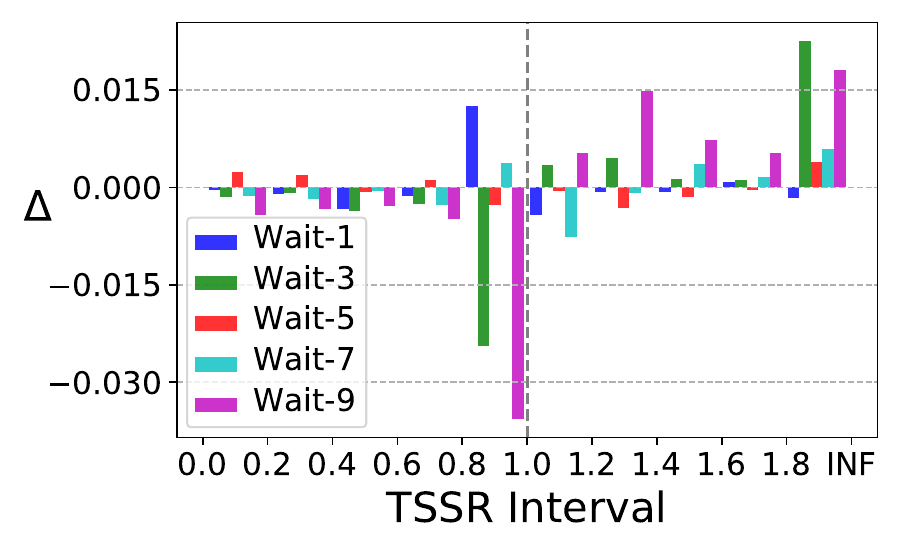}
  \captionof{figure}{Word Frequency Rate Change ($\Delta$) in different TSSR intervals with scheduled sampling training compared to the Baselines.}
  \label{fig:word_freq_rate_change_in_tssr}
\end{figure}

\begin{figure}[H]
  \centering
  \includegraphics[width=2.0in]{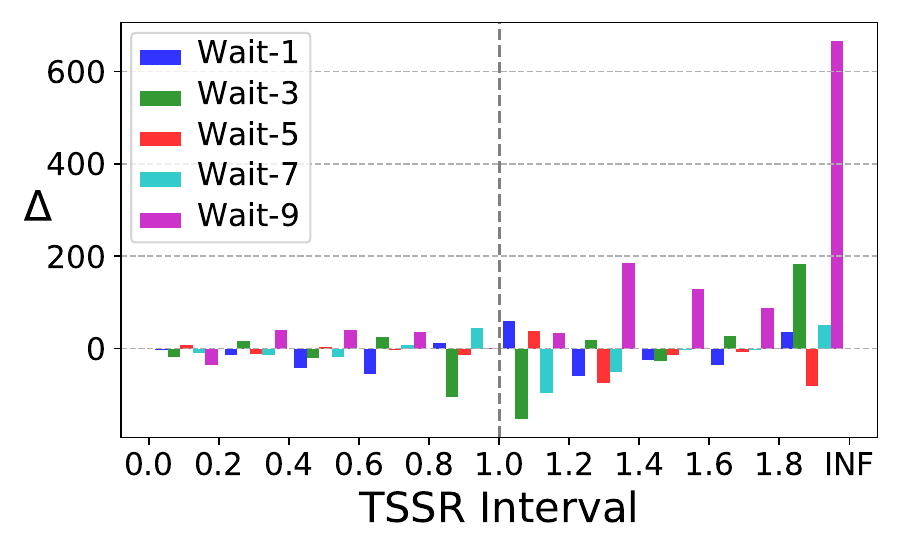}
  \captionof{figure}{Hallucination Frequency Change ($\Delta$) in different TSSR intervals with scheduled sampling training compared to the Baselines.}
  \label{fig:hall_freq_change_in_tssr}
\end{figure}

\begin{table}[H]\centering 
  \setlength{\tabcolsep}{0.3em}
  \renewcommand{\arraystretch}{0.6}
  \resizebox{\linewidth}{!}{
  \begin{tabular}{lrrrrrr}
  \toprule
  \multicolumn{2}{l}{ } &  $k$=1  & $k$=3  & $k$=5 & $k$=7 & $k$=9  \\ 
  \midrule
  \multirow{2}{*}{Baselines} & BLEU $\uparrow$ &12.33 & 15.39 & 16.26 & 16.66 & 16.66  \\
     & \textrm{HR} $\% \downarrow$ &{33.96} & {25.31} & {23.22} & {21.84} & {20.73}  \\
  \midrule
  Scheduled-  & BLEU $\uparrow$ & 12.42 & 15.51 & 16.43 & 16.61 & 17.03 \\
  Sampling & \textrm{HR} $\% \downarrow$ & {33.69} & {25.29} & {22.68} & {21.61} & {23.50}  \\
  \bottomrule 
  \end{tabular}
  }
  \caption{BLEU scores and \textrm{HR} of wait-$k$ models.}
  \label{tab:bleu_and_hr}
\end{table}

\section{Alignment Error Rate of Awesome-Align}

\begin{table}[H]
  \centering
  \begin{adjustbox}{width=0.7\columnwidth}
  \begin{tabular}{c|c}
    \toprule
    \textrm{Alignment Error Rate}  & 7.30 $\%$ \\
    \textrm{Precision } & 0.950 \\
    \textrm{Recall} & 0.885 \\
    \bottomrule
  \end{tabular}
  \end{adjustbox}
  \caption{The alignment error rate, precision, and recall of hallucination detection using Awesome-align, with human annotations as the ground truth.}
  \label{tab:dhr_in_waitk}
\end{table}

We report the alignment error rate as well as the precision and recall of hallucination detection using Awesome-align. Based on the precision and recall results, we believe that the automatic word alignment is suitable for detecting hallucinated words.

\end{CJK}
\end{document}